%% file: main.tex
\documentclass[10pt,twocolumn,letterpaper]{article}

\usepackage[pagenumbers]{iccv} 
\input{preamble}

\definecolor{iccvblue}{rgb}{0.21,0.49,0.74}
\usepackage[pagebackref,breaklinks,colorlinks,allcolors=iccvblue]{hyperref}

\title{Vision-Speech Models: Teaching Speech Models to  Converse about  Images}

\author{
Amélie Royer$^*$\\
Kyutai\\
{\tt\small amelie@kyutai.org}
\and
Moritz Böhle$^*$\\
Kyutai\\
{\tt\small moritz@kyutai.org}
\and
Gabriel de Marmiesse\\
Kyutai\\
{\tt\small gabriel@kyutai.org}
\and
Laurent Mazaré\\
Kyutai\\
{\tt\small laurent@kyutai.org}
\and
Neil Zeghidour\\
Kyutai\\
{\tt\small neil@kyutai.org}
\and
Alexandre Défossez\\
Kyutai\\
{\tt\small alex@kyutai.org}
\and
Patrick Pérez\\
Kyutai\\
{\tt\small patrick@kyutai.org}
}

\begin{document}
\maketitle

\input{sections/0-abstract}    
\input{sections/1-intro}
\input{sections/2-related}
\input{sections/3-methods}
\input{sections/4-experiments}

\input{sections/conclusions}
{
    \small
    \bibliographystyle{ieeenat_fullname}
    \bibliography{main}
}
\input{sections/5-appendix}

\end{document}

%% file: preamble.tex
\usepackage{pifont}

\usepackage{enumitem}
\usepackage{tcolorbox}
\usepackage{listings}
\usepackage{booktabs}

\usepackage{fontawesome}
\usepackage{subcaption}
\usepackage{tikz}
\usepackage[normalem]{ulem}
\usepackage{multirow}

\usepackage{changepage}
\usepackage[export]{adjustbox}

\newcommand{\myparagraph}[1]{\vspace{.25em}{\noindent\textbf{#1}}}

\newcommand{\myfigref}[1]{\hyperref[#1]{Figure \ref{#1}}}
\newcommand{\mysecref}[1]{\hyperref[#1]{Section \ref{#1}}}
\newcommand{\mytblref}[1]{\hyperref[#1]{Table \ref{#1}}}

\newcommand\ocrvqa{OCR-VQA\xspace}
\newcommand\tallyqa{TallyQA\xspace}
\newcommand\pixmo{PixMo\xspace}
\newcommand\docci{DOCCI\xspace}
\newcommand\pixelprose{PixelProse\xspace}
\newcommand\docvqa{DocVQA\xspace}
\newcommand\vqa{VQAv2\xspace}
\newcommand\mosnet{MOSNet\xspace}
\newcommand\coco{COCO\xspace}

\newcommand\cider{CIDEr\xspace}
\newcommand\moshivis{MoshiVis\xspace}
\newcommand\githublink{\href{https://github.com/kyutai-labs/moshivis}{github.com/kyutai-labs/moshivis}\xspace}

\newcommand{\myeq}{\mkern1.25mu{=}\mkern1.25mu}
\newcommand{\mygt}{\mkern1.25mu{>}\mkern1.25mu}

\definecolor{ocrvqa}{RGB}{56, 192, 243}
\definecolor{mosnet}{RGB}{200, 200, 5}
\definecolor{coco}{RGB}{112, 224, 65}
\definecolor{vqav2}{RGB}{190, 190, 40}
\definecolor{texteval}{RGB}{145, 145, 145}
\definecolor{camodule}{RGB}{255, 184, 83}
\definecolor{audiotoks}{RGB}{0, 132, 219}
\definecolor{texttoks}{RGB}{23, 171, 0}

\newcommand{\coloredbox}[1]{\raisebox{0.55ex}{\fcolorbox{black}{#1}{\rule{0pt}{.05em}\rule{.05em}{0pt}}}}

\usepackage{caption}

\newcounter{tboxcounter}
\newcommand{\tboxcaption}[1]{\refstepcounter{tboxcounter}\vspace{-\medskipamount}\vspace{1em}\par\noindent\centerline{\parbox{.925\linewidth}{\raggedright    \textbf{Instruction \thetboxcounter} #1}}}
\usepackage[pagebackref,breaklinks,colorlinks,allcolors=iccvblue]{hyperref}
\usepackage[capitalize]{cleveref}
\crefname{tboxcounter}{Instruction}{Instructions} \Crefname{tboxcounter}{Instruction}{Instructions} 

\let\svthefootnote\thefootnote
\newcommand\freefootnote[1]{%
  \let\thefootnote\relax%
  \footnotetext{#1}%
  \let\thefootnote\svthefootnote%
}

%% file: sections/0-abstract.tex
\begin{abstract}
The recent successes of Vision-Language models raise the question of how to equivalently imbue a pretrained speech model with visual understanding, an important milestone towards building a multimodal speech model able to freely converse about images. 
Building such a conversational Vision-Speech model brings its unique challenges: \textbf{(i)} paired image-speech datasets are much scarcer
 than their image-text counterparts, \textbf{(ii)} ensuring real-time latency at inference is crucial thus bringing compute and memory constraints, and \textbf{(iii)} the model should preserve prosodic features (\eg, speaker tone) which cannot be inferred from text alone.  
In this work, we introduce \moshivis, augmenting a recent dialogue speech LLM, Moshi, with visual inputs through lightweight adaptation modules. An additional dynamic gating mechanism enables the model to more easily switch between the visual inputs and unrelated conversation topics. 
To reduce training costs, we design a simple one-stage, parameter-efficient fine-tuning pipeline  in which we leverage a mixture of image-text (\ie, ``speechless'') and image-speech samples.
We evaluate the model on downstream visual understanding tasks with both audio and text prompts, and report qualitative samples of interactions with \moshivis.
Our inference code, the image-speech data used for audio evaluation, as well as additional information are available at \githublink. \freefootnote{$^*$\,Equal contribution.}
\end{abstract}

%% file: sections/1-intro.tex
\section{Introduction}
\label{sec:intro}

\input{resources/latex/figures/main}

Vision Language Models (VLMs) have recently gained increasing attention, \eg~\cite{agrawal2024pixtral,molmo2024,beyer2024paligemma,steiner2024paligemma,Qwen-VL,internlmxcomposer,liu2023llava}, showcasing strong capabilities across a variety of visual understanding tasks such as question answering, image captioning or complex reasoning over visual inputs. 
A core challenge of training VLMs, or multimodal models in general, is to build well-aligned embeddings of the different input modalities.
To this end, the VLM research community has built up vast datasets of paired image and text data over the years, covering many vision  understanding tasks~\cite{gadre2023datacomp,molmo2024,schumann2022laion5b,li2024recaption,kembhavi2016ai2d,thecauldron}.
In comparison, such public datasets are very rare in the speech domain, and often limited to captions~\cite{havard2017speech,spokencoco}. This lack of data is particularly apparent when considering the challenge of building open-source multimodal models able to naturally talk about an image as well as other general topics,  even though such models are starting to appear in the commercial space \cite{projectastra,openaiVision}. 

In this work, we aim to effectively integrate vision capabilities into a conversational speech LLM.
As our backbone, we use Moshi~\cite{kyutai2024moshi}, a recent open-weight speech LLM able to dialogue with the user in real time and in full-duplex, \ie it is able to listen and speak at any time and does not need to be signalled when to talk. 
Drawing inspiration from  VLMs, we aim to adapt Moshi into a \textbf{Vision-Speech Model} (VSM) with the same dialoguing abilities.  
We identify three key challenges specific to building a VSM able to hold natural conversations about visual inputs:
\textbf{(i)} overcome the above-mentioned scarcity of image-aligned speech data, and avoid blowing up the complexity of the training pipeline as we are now dealing with three modalities---vision, language and audio, \textbf{(ii)} comply with compute and memory constraints in order to hold real-time conversations at inference, and
\textbf{(iii)} maintain the original conversational abilities of the backbone dialogue model, \ie, preserve audio quality as well as prosodic features, and enable seamless
switching between image-related and general conversation topics.

We address each of these challenges as follows:
\textbf{First,} we show that we can adapt the underlying pretrained speech transformer  to image inputs using image-text datasets without audio supervision (``speechless'' datasets), in combination with a small percentage of speech samples. 
Specifically, we exploit the fact that Moshi \textit{jointly} predicts text and audio tokens in a temporally aligned manner. While the produced text tokens differ in distribution from standard language, we find that this  form of weak  supervision 
still allows for information transfer from the text to speech modality, despite being out-of-distribution for  the backbone model. 
\textbf{Second,} inspired by recent work on perceptual augmentations~\cite{shukor2023epalm,Vallaeys2024ImprovedBF}, we inject the visual inputs into the speech LLM backbone through lightweight adaptation modules based on cross-attention. At inference, the keys and values of the attention mechanism can be efficiently cached and thus only need to be computed once for every image. 
\textbf{Third,} to ensure that the base model is still able to discuss general conversation topics other than the input image, 
we design these cross-attention layers to be able to selectively gate visual inputs based on the conversational context.
In addition, expanding on previous work on visual dialogues~\cite{das2017visual,wen2023infovisdial}, we design a fully synthetic data pipeline to generate short realistic conversations about images, allowing the model to go beyond the usual setting of ``one question - one answer'' assumed in most VLM benchmarks. 

An overview of our proposed model, \textbf{MoshiVis} is given in \hyperref[fig:overview]{Figure \ref{fig:overview}}: We augment a pretrained spoken dialogue model, Moshi, with lightweight adaption modules and a simple training pipeline leveraging both image-text and image-speech data. 
To assess the model's visual and conversational abilities, we first evaluate MoshiVis's visual understanding on downstream tasks commonly used in the VLM literature,  such as captioning or visual question answering, in \textit{both} the text and audio realms. To foster further research on Vision-Speech models, we  release the audio datasets we use for benchmarking. 
We then evaluate the model's ability to switch contexts by measuring how its text reasoning and visual understanding abilities are affected when adding an irrelevant conversation snippet as a prefixing context. 
Finally, we provide qualitative samples of dialogues with MoshiVis to highlight its conversation abilities and low latency: For instance, on a L4  GPU,  we find that MoshiVis only increases latency by 7ms per inference step compared to the base model Moshi, preserving real-time interaction.

\textbf{In short}, our contributions are: \textbf{(i)} a simple one-stage training recipe that leverages ``speechless'' data in combination with speech samples, tapping into the large amounts of pre-existing vision-language datasets; \textbf{(ii)} a lightweight gating mechanism to facilitate context switches in conversations, in particular between image-related and non-relevant content; \textbf{(iii)} a synthetic data pipeline for generating realistic visual dialogues. To facilitate reproducibility, we release our inference code as well as the image-speech benchmarks used to evaluate the model in audio form. 

%% file: resources/latex/figures/main.tex
\begin{figure*}[t]
    \centering
        \includegraphics[width=.82\linewidth]{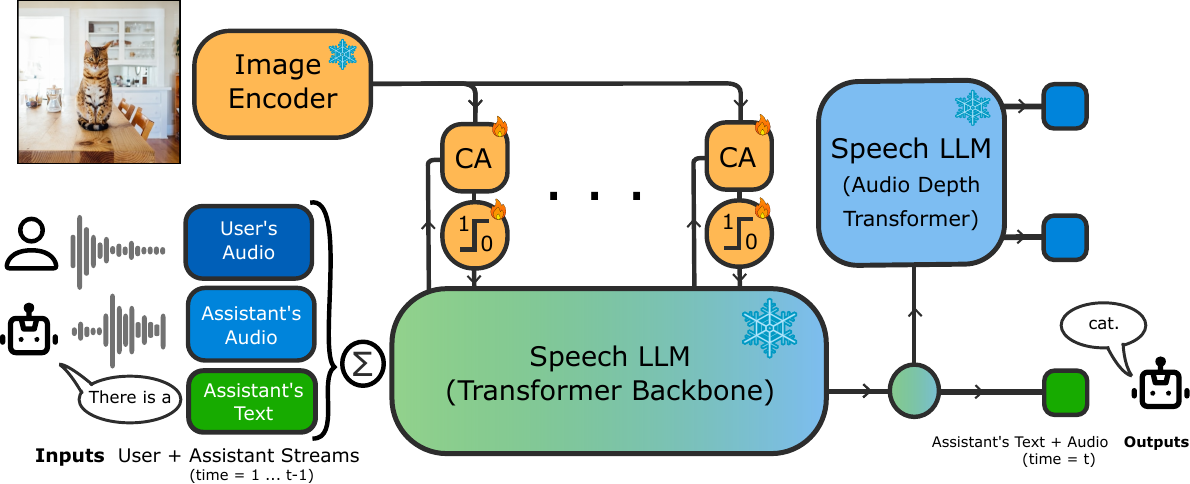}
        \vspace{-0.05cm}
        \caption{\textbf{MoshiVis} is a Vision-Speech model (\textbf{VSM}) able to hold  full-duplex real-time conversations about an image, and trained with a  light data- and compute- budget. 
        For image representations, we use  off-the-shelf transformer-based image encoders  from the PaliGemma family~\cite{beyer2024paligemma}. 
        For the speech modelling part, we rely on Moshi~\cite{kyutai2024moshi}, a recent speech LLM  which \textit{jointly} outputs text and audio tokens in real-time, allowing for full-duplex conversations. 
        At its core, Moshi consists of a standard 7B decoder-only transformer taking as inputs   \textit{speech tokens} (which are the sums of temporally aligned {\color[rgb]{0.1, .6, .1}text} tokens and {\color[rgb]{0.1, .2, .7}audio} tokens extracted from the assistant's and user's streams), rather than only text like a standard LLM. The output of the transformer is then separately decoded in a text token, as well as passed through a small \textit{depth transformer} which auto-regressively produces a hierarchy of audio codebooks, then decoded into audio frames. 
        First, (\hyperref[sec:align]{ Sec. \ref{sec:align}}), we detail how we augment the speech LLM's transformer with lightweight visual adaptation modules 
        through cross-attention (CA). We then describe our one-stage finetuning pipeline  for these modules:
        We use a mixture of \textbf{(i)} (\hyperref[sec:train]{Sec. \ref{sec:train}}) image+text only data (\textit{``speechless"} data), which, despite incurring a distribution shift due to the lack of audio supervision, allows us to leverage the large body of existing Vision-Language datasets, and \textbf{(ii)} (\hyperref[sec:data]{Sec.  \ref{sec:data}})  synthetic spoken visual dialogues which we design to mimic realistic discussions about images.              \vspace{-0.31cm}
    } 
    \label{fig:overview}
\end{figure*}

%% file: sections/2-related.tex
\section{Related Work}
\label{sec:related}

\myparagraph{Vision Language Models.} VLMs transfer to visual inputs the strong reasoning abilities of LLMs, to achieve
complex visual understanding~\cite{beyer2024paligemma,steiner2024paligemma,agrawal2024pixtral,lee2025vhelm,internlmxcomposer,Qwen-VL,molmo2024}.
By combining an LLM and an image encoder, they show remarkable results on various visual understanding tasks, such as captioning, question answering or optical character recognition. While early-stage joint pre-training is a popular technique to train  VLMs~\cite{beyer2024paligemma,agrawal2024pixtral,molmo2024,internlmxcomposer,Qwen-VL}, it generally requires large amounts of image-text data and well-tuned multi-stage training pipelines. Expanding this approach to the additional audio modality, as done in \cite{miniomni2} for instance, comes with costly training data and compute requirements. 
Instead, we draw inspiration from VLM ``perceptual augmentations'', which have proven data- and parameter-efficient while still achieving strong visual capabilities~\cite{Vallaeys2024ImprovedBF,manas2023mapl,shukor2023epalm,alayrac2022flamingo}.
Such methods typically first project the image tokens to a more amenable embedding space, then inject these tokens in the text token flow via prefixing or cross-attention. 
Similarly, in this work, we introduce 
adaption modules based on gated cross-attention to adapt a speech model into a VSM.  
This choice is primarily motivated for practical reasons: Direct insertion of the image tokens effectively takes space in the context window of the model, thus limiting the length of the conversation which is often bottlenecked by the size of the KV cache at inference.

\myparagraph{Speech Modelling and Visual Inputs.} 
A straightforward way to augment a VLM with speech 
is to use ad-hoc text-speech conversion:  an input module transcribing the input speech, and an output text-to-speech (TTS) module producing speech from the LLM outputs~\cite{hfvoicechat}. 
However, it is well known in the speech modelling literature~\cite{nguyen2024spiritlminterleavedspokenwritten,fang-etal-2024-llama-omni,kyutai2024moshi} that this cascaded setup has severe flaws: 
it causes noticeable latency, loses prosodic information such as the user's tone or emotion because of the input speech transcription, and it imposes separated speaker turns.
Another alternative would be to include the audio modality directly at the pretraining stage of a VLM, building for instance on image-speech encoders such as SpeechCLIP~\cite{Shih2022SpeechCLIPIS}. While this joint training  approach is being successfully explored in projects such as Mini-Omni2~\cite{miniomni2} or AnyGPT~\cite{zhan2024anygpt}, aiming to reproduce the abilities of closed commercial multimodal assistants such as GPT-4o, it requires a carefully crafted multi-stage training pipeline and datasets selection to balance all modalities across the stages.
Instead, in this work, we  leverage a pretrained speech LLM, \ie, a voice model with strong built-in conversational abilities, and we expand on VLM perceptual augmentation techniques to propose a simple training pipeline for turning the speech model into a VSM. 
Specifically, we rely on Moshi~\cite{kyutai2024moshi}, a recent open-weight Speech LLM which \textit{jointly} produces text and audio. 
As we will show, the presence of this text stream, despite having a distribution different from standard text, provides a strong basis for leveraging VLM techniques and thus to adapt Moshi into a VSM. Moreover, as Moshi was designed as a real-time conversational model, 
it proves itself to be a good starting point for designing a real-time conversational VSM. 

\newpage 
\myparagraph{Towards Multimodal Dialogue Models.} 
Extending multimodal models from the standard ``one question - one answer'' paradigm to more natural multi-turn conversations is a challenging task, both from the training and evaluation perspective. 
In the language domain, early work on visual dialogues~\cite{das2017visual,wen2023infovisdial} introduced the task of answering a sequence of around 10 questions about an image. 
Expanding on this, our work also aims to further explore how one can go from a VSM to a model able to dialogue about an image at will. 
In particular, we investigate the model's ability to switch context between image-relevant and more general conversation topics, inspired from task switching analysis in LLMs~\cite{gupta-etal-2024-llm}. We also design a synthetic data pipeline modelling realistic conversations about images (\eg questions with different levels of details, misleading questions, \etc).

%% file: sections/3-methods.tex
\section{Design and Training of MoshiVis}

We now describe how we augment a speech LLM such as Moshi~\cite{kyutai2024moshi} to handle visual inputs, 
while maintaining its conversational capabilities and real-time latency.
\noindent In \hyperref[sec:align]{Section \ref{sec:align}}, we describe how we inject visual information into the stream of speech tokens, as shown  in \hyperref[fig:overview]{Figure \ref{fig:overview}}. 
In \hyperref[sec:train]{Section \ref{sec:train}}, we discuss how we leverage standard image-text data, allowing us to directly tap into the large body of vision-language understanding datasets, instead of having to collect large amounts of dedicated image-speech data. 
Nevertheless, for training a visual \textit{conversational} model we still lack adequate, freely accessible dialogue datasets. To remedy this, we introduce in \hyperref[sec:data]{Section \ref{sec:data}} a synthetic data generation pipeline for producing  spoken visual dialogues. 

\subsection{Image-Speech Adaptation}
\label{sec:align}
As the core backbone of the proposed architecture, we use Moshi~\cite{kyutai2024moshi}, a recent end-to-end speech dialogue model which jointly predicts text and audio tokens in real time. 
Our aim is to augment this  backbone to interpret visual inputs given by a pretrained image encoder,  such that it preserves the low inference latency required for real-time conversations, and while keeping a reasonable training budget. 
Our proposed pipeline is agnostic to the specific choice of the image encoder, as long as it outputs a tokenized representation of the image. In practice, we use off-the-shelf state-of-the-art image encoders from  PaliGemma~\cite{beyer2024paligemma}. 

\myparagraph{Preliminaries: Speech LLMs.}
As shown in \myfigref{fig:overview}, during its forward pass, Moshi first encodes the input dialogue into multiple temporally-aligned streams of tokens: A text  stream capturing the assistant's speech content, and two audio streams, one for the assistant and the user respectively. 
These tokens are then summed to form a single stream of tokens, the \textit{speech tokens},  which are then fed to a transformer. 
The output sequence of speech embeddings are finally decoded back into separate  text and audio tokens with a lightweight \textit{depth transformer}. 
For further details about Moshi, in particular on  audio processing, please refer to the original paper~\cite{kyutai2024moshi}.
Importantly, while speech tokens contain information from the text stream, their distribution  differ from standard text used in language modelling: \textbf{(i)} they are summed with  audio tokens hence contain  additional non-semantic acoustic information, and \textbf{(ii)} the underlying text stream processed by Moshi contains many additional padding tokens to preserve the temporal alignment between text and speech. 
Nevertheless, the core backbone of Moshi can be seen as a standard transformer acting on \textit{speech} tokens, which we aim to further augment to be able to process visual inputs.

\input{resources/latex/figures/cross_attention}

\myparagraph{Gated Cross-Attention.} To fuse image information into the stream of speech tokens, we introduce a cross-attention layer in each transformer block, as illustrated in \hyperref[fig:gatedxa]{Figure \ref{fig:gatedxa}}. The cross-attention takes as queries the tokens output by the self-attention layer, and uses the image embeddings as keys and values. The output is then used to compute a residual update of the speech tokens.
However,  introducing this additional source of information may be detrimental to the model's initial conversational abilities, in particular its ability to switch context (see ablation experiments in \hyperref[sec:context_switching]{Section \ref{sec:context_switching}}).    
To promote context switching abilities, we further modulate the output of the cross-attention module with a self-gating mechanism. Intuitively, a gate output of zero would turn off the image information and exactly recover the base model behaviour, while higher values facilitate the flow of image information. Specifically, the gate is a small 2-layer MLP with a hidden size reduction factor of 1/8, followed by a sigmoid activation. During training, we do not supervise the gate's outputs and instead let it implicitly learn an image relevance score for each token. 

\myparagraph{Compute Efficiency.}
At inference, as the image tokens are independent of the speech tokens, we can precompute and cache their KV projections once at the beginning of the temporal stream. 
In addition, we use the same cross-attention QKV projection weights in every layer of the transformer, which lessens the memory cost of the aforementioned cache, and we find that it does not significantly impact performance (see  \hyperref[sec:ablations]{Section \ref{sec:ablations}}). 
As for training cost, note that in all of our experiments we keep the weights of the image embedder and the speech transformer frozen. 
We find that this has two positive effects:  \textbf{(i)}
It enables a lightweight training pipeline that is accessible to a wider audience for downstream task tuning; in total, we only train the adaptation modules which amounts to a total of 206M trainable parameters,  and \textbf{(ii)} it avoids a potential degradation of the backbone speech transformer's ability to converse about general topics other than the input image. In particular,  our model exactly recovers the backbone speech transformer when the gates' outputs are all zeros.

\input{resources/latex/figures/forwardpass}

\subsection{Leveraging ``Speechless'' Datasets for Training}
\label{sec:train}

While image-text datasets have flourished in recent years, equivalent datasets in speech form are scarcely available and mainly consist of  transcripts of COCO-Captions~\cite{Hsu2020TextFreeIS,havard2017speech,spokencoco}. 
In \hyperref[sec:data]{Section \ref{sec:data}}, we further expand this line of work by introducing a pipeline to  synthetically generate realistic spoken visual dialogues. 
However, the cost of generating such data (and associated training time) to cover all aspects of visual understanding tasks, as well as speech properties (\eg, variety of  prosody, emotions, speaker interruptions, \etc) would quickly blow up.  
As an alternative, we aim to tap into existing image-text data which already covers a wide variety of visual understanding tasks. To that end, a key observation is that the backbone speech model explicitly predicts (and takes as input) a stream of text tokens.
However, as illustrated in \hyperref[fig:forwardpass]{Figure \ref{fig:forwardpass}}, this stream of text tokens contains many occurrences of padding tokens---required to temporally align text and audio streams--- thus does not follow the same distribution as standard text (as would, \eg, a direct transcript of the audio).
Nevertheless, we hypothesize it is possible to train our adaptation modules on ``speechless'' data, even though this incurs a distribution shift as \textbf{(i)} the model  expects summed audio and text tokens, and \textbf{(ii)}  speechless samples do not provide any supervisory signal to the audio codebooks output by the model.

 To alleviate this, we train \moshivis with mixed supervision: Each batch of data is composed of $p_{\text{audio}}\%$  speech samples with audio streams,  and $\left(100 - p_{\text{audio}}\right)\%$\ speechless samples. 
  As shown in \hyperref[fig:forwardpass]{Figure \ref{fig:forwardpass}}, speech samples are in-distribution with respect to the base Moshi model: the corresponding text stream only contains the text of the assistant, while the user speech is only present as audio (as the corresponding text would not be available at inference). 
  In contrast, speechless samples contain the whole transcript in text (including the user's questions)---as such, they differ significantly from the speech inputs: their  stream of text tokens does not include any alignment padding tokens, the user's  input is given in text instead of audio, and finally, they do not contain any audio information.
  Interestingly, we nonetheless find that even a few audio samples in the batch are sufficient for the model to learn from the text-only signal while preserving coherent speech in the output, as we show later in experiments (\hyperref[sec:text_transfer]{Section \ref{sec:text_transfer}}). 
Importantly, this means we can now finetune the model on specialized downstream vision tasks using readily available image-text data, with little audio supervision.  

Next, we discuss the case of visual dialogue datasets, which are scarcely available in text and inexistent in speech.

\subsection{Generating Synthetic Visual Dialogues}
\label{sec:data}
Visual dialogue datasets only exist in text form~\cite{das2017visual,wen2023infovisdial} and  often  consist of fixed-length sequences of short question-answer pairs. 
To promote more natural conversational flow, we design a synthetic data  pipeline for spoken visual dialogues, which we use to train the final dialogue model. 

\myparagraph{Spoken Visual Dialogue Generation. }
Our first step is to generate realistic conversation about images in text form. For this, we prompt two separate Mistral-Nemo~\cite{mistralnemo} models in turns, one with the goal of asking questions (the \textit{user}) the other to answer them (the \textit{assistant}). Both LLMs are also fed with the same text caption of an image to use as support for their respective roles.
To start off the dialogue, we prompt the user to ask a general question about the image (\eg ``\textit{what's in the image?}") and for the assistant to give a global description in a few sentences. The initial question prompt is also designed to broadly cover different question lengths, conversation tones and vocabulary. 
The models then continue the dialogue for 8 to 16 turns (a turn being a question-answer pair), while being prompted with a randomly selected instruction at each turn. We design several instructions, each capturing  a different type of conversation about an image, such as general questions about the image content, about fine-grained details (object locations and their properties), as well as misleading questions (\eg about objects not present in the image). 

All prompts used for data generation are given in \hyperref[app:promptengineering]{Appendix \ref{app:promptengineering}}. 
Once the text dialogues are generated, we convert them to speech using the same text-to-speech model as in \cite{kyutai2024moshi}, ensuring a consistent assistant voice across samples. 

\myparagraph{Data Augmentation.}
To further enhance the model's ability to switch topics during a conversation, we also generate a set of generic spoken dialogues, not related to any image, following the synthetic data procedure described in \cite{kyutai2024moshi}.  At training time,  each visual dialogue has a $p_{\text{concat}}$ chance of being concatenated on-the-fly with a prefix and suffix conversation, randomly sampled  from this set of unrelated dialogues. In addition, we randomly sample and trim  the length of each of the three dialogues being concatenated (\ie, the prefix, suffix and visual dialogue).

%% file: resources/latex/figures/cross_attention.tex
\begin{figure}[t]
    \centering
    \includegraphics[width=1\linewidth]{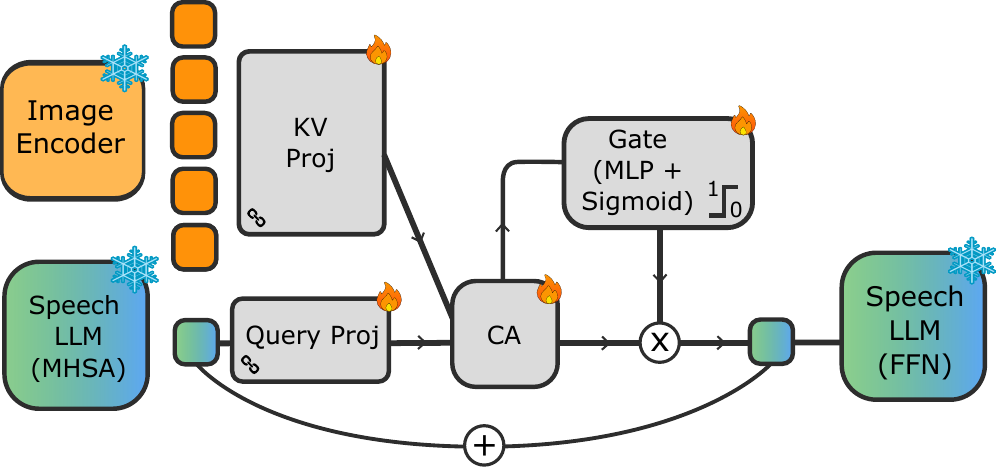}
    \caption{\textbf{Adaptation modules.} The image tokens are injected into the current speech token via residual cross-attention (CA) layers,  placed between the multi-head self attention (MHSA) and the feedforward network (FFN) in every transformer block. As the cross-attention's QKV projections are shared across layers $\left(\text{\faChain{}}\right)$, at inference, we only need to compute the keys and values once per image, thus reducing the memory cost needed to store the image embeddings. To enable more context switch, we modulate the output of the cross-attention with a binary gate. The resulting output is fed back into the speech token stream as a residual.  
    \vspace{-0.31cm}
    }
    \label{fig:gatedxa}
\end{figure}

%% file: resources/latex/figures/forwardpass.tex
\begin{figure}
    \centering
    \includegraphics[width=0.95\linewidth]{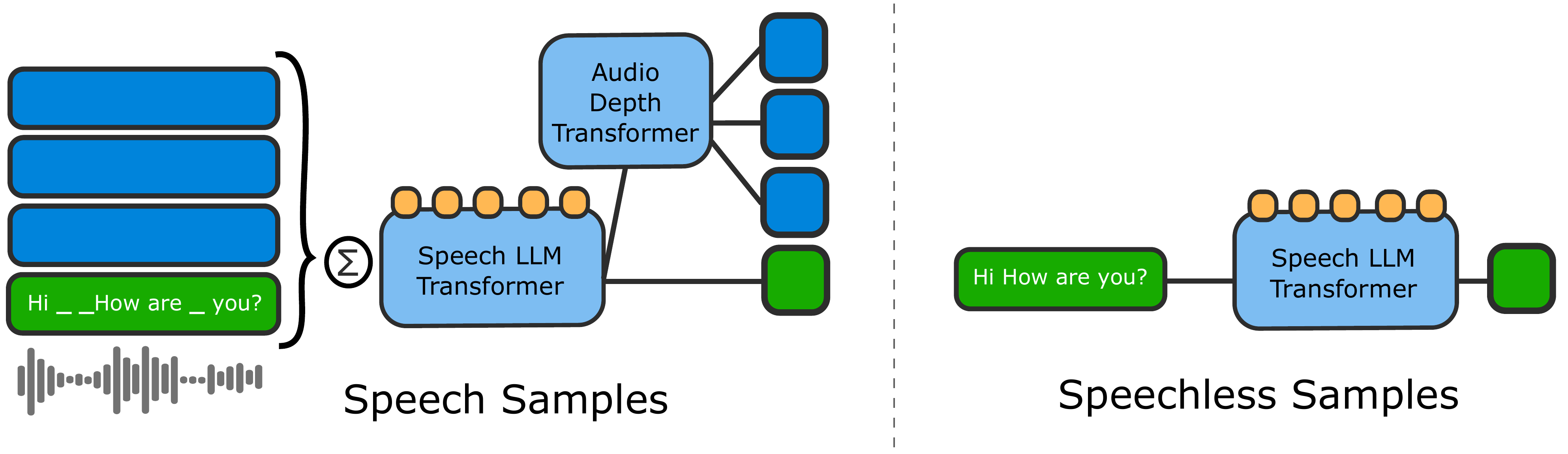}
    \caption{\textbf{\moshivis forward pass during mixed data training}. 
        \textit{Speech samples} are composed of the user's and assistant's audio streams (\coloredbox{audiotoks}) and a text stream (\coloredbox{texttoks}) (only for the assistant)  containing extra padding tokens (\_) to maintain the temporal alignment with speech. The input streams are summed and  passed to the transformer. The output audio streams are auto-regressively decoded by a small transformer (Audio Depth Transformer). In practice, \textit{we only train the first two audio streams for speech samples}. 
    This allows for faster training as we need fewer parallel calls to the depth transformer. 
        In contrast, \textit{speechless samples} only contain standard text; in this case, \moshivis acts as a standard transformer augmented with additional adaptation modules (\coloredbox{camodule}).     \vspace{-0.25cm}}
    \label{fig:forwardpass}
\end{figure}

%% file: sections/4-experiments.tex
\section{Experiments}

\input{resources/latex/figures/text-speech-ratio} 

In this section, we discuss the performance of MoshiVis in practice. 
First, in \hyperref[sec:text_transfer]{Section \ref{sec:text_transfer}} we evaluate its downstream accuracy on classical vision tasks including generic image understanding (captioning, question answering) and more specialized tasks (text reading). 
In particular, we evaluate each task in both the text and audio domains, and carefully analyse how the proportion of audio data available at training affects visual understanding as well as audio quality. 
Secondly, we address our initial target task and discuss the model's ability to hold a spoken conversation about visual inputs. In \hyperref[sec:context_switching]{Section \ref{sec:context_switching}}, we measure the model's ability to switch between different contexts in a single conversation, \ie, going from talking about the image to an entirely different topic, and vice-versa, and how this behaviour is affected by the gating mechanism. 
Finally, in \hyperref[sec:mixmodel]{Section \ref{sec:mixmodel}}, we discuss practical usage of \moshivis ``in-the-wild'', such as real-time inference latencies and qualitative samples.

\subsection{Vision-Speech Benchmark}
\label{sec:text_transfer}
\input{sections/4-1-benchmarks}

\subsection{Evaluating Robustness to Context Switches}
\label{sec:context_switching}
\input{sections/4-2-contextswitch}

\subsection{MoshiVis in-the-wild}
\label{sec:mixmodel}
\input{sections/4-3-realworld}

%% file: resources/latex/figures/text-speech-ratio.tex
\begin{figure*}[tbh]
    \centering
    \begin{subfigure}[c]{.325\textwidth}{
    \includegraphics[width=\linewidth]{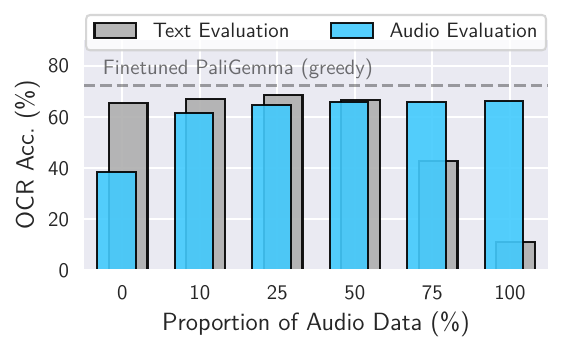}
    \vspace{-0.5cm}
    }
    \caption{\textbf{\ocrvqa}}
    \end{subfigure}
        \begin{subfigure}[c]{.325\textwidth}{
    \includegraphics[width=\linewidth]{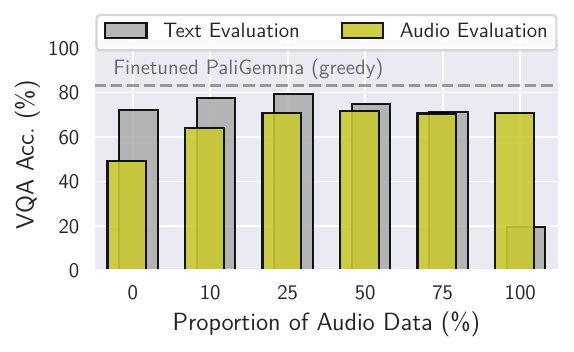}
    \vspace{-0.5cm}
    }
    \caption{\textbf{\vqa}}
    \end{subfigure}
        \begin{subfigure}[c]{.325\textwidth}{
    \includegraphics[width=\linewidth]{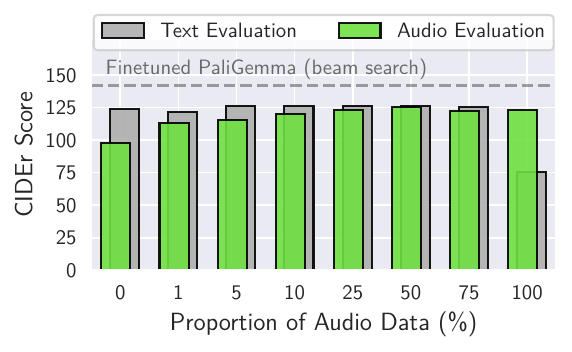}
    \vspace{-0.5cm}
    }
    \caption{\textbf{COCO}}
    \end{subfigure}
        \vspace{-0.08cm}
    \caption{\textbf{Training MoshiVis with different amounts of audio data} on \textbf{a)} \ocrvqa, \textbf{b)} \vqa, and \textbf{c)} \coco. In particular, we show the scores obtained by the model when prompting it either with text (\coloredbox{texteval}) or audio (\coloredbox{ocrvqa} \coloredbox{vqav2} \coloredbox{coco}) and using greedy decoding. Note that even when training \textit{with no audio data at all}, the cross-attention mechanism enables the speech model to obtain results substantially above chance on all benchmarks. While this can come at the cost of audio quality, we find that adding as little as 1\% of audio data effectively recovers the base model's audio quality  (\mytblref{tab:mosnet_results}). For reference, we also report the results of the fine-tuned PaliGemma (stage 3 of~\cite{beyer2024paligemma}), as we use the same image encoder as a starting point for fine-tuning; \textit{note that in contrast to PaliGemma, we keep the image encoder and LLM frozen}.         \vspace{-0.30cm}
    }
    \label{fig:text_speech_results}
\end{figure*}

%% file: sections/4-1-benchmarks.tex
As discussed in \hyperref[sec:train]{Section \ref{sec:train}}, we train \moshivis with mixed data, each  batch containing a proportion $p_{\text{audio}}$ of speech samples.
In this section, we assess whether training  the model with speechless data still translates to actual vision understanding when queried via audio/speech, and, in  particular,  how $p_{\text{audio}}$ affects this performance. Note that in this section \textit{we do not use any of the synthetic visual dialogues  introduced in \hyperref[sec:data]{Section \ref{sec:data}}}. Instead, we focus on downstream performance on specific vision-language benchmarks.

\myparagraph{Experimental Setup.}
\label{sec:text_transfer:setup}
As backbones, we use Moshi~\cite{kyutai2024moshi} for the speech modality ($\sim$ 7B parameters), and the ``stage 2'' vision encoder of PaliGemma~\cite{beyer2024paligemma}  ($\sim 400$M parameters) for images. Both backbones are kept frozen during training, and  we only train the adaptation modules ($\sim 206$M parameters). 
We employ  benchmarks covering a range of visual understanding tasks: \coco~\cite{lin2014mscoco} (image captioning), \ocrvqa~\cite{mishraICDAR19} (text recognition), 
and \vqa~\cite{goyal2017vqav2} (question answering).  
As we primarily want to evaluate our vision-speech model in the audio domain, we also convert these datasets to speech using the same text-to-speech model as in our synthetic data generation pipeline; the resulting image-speech datasets for evaluation are available on our project page, see \githublink. Note that prompting the model with speech instead of text introduces interesting challenges, for instance due to certain benchmarks being sensitive to formatting such as text punctuations, which are not necessarily transcribed in speech data. We discuss these in more detail in \hyperref[app:audioeval]{Appendix \ref{app:audioeval}}.

\myparagraph{Main Results.} In \hyperref[fig:text_speech_results]{Figure \ref{fig:text_speech_results}}, we report downstream performance after training the adaptation modules in MoshiVis on three separate tasks (\ocrvqa, \vqa, and \coco), while varying the proportion of samples with audio in the batch $p_{\text{audio}}\,{\in}\,\{0\%, 10\%, 25\%, 50\%, 75\%, 100\%\}$.
We evaluate the model by prompting it in both text and audio form. In both scenarios, we directly use the text tokens generated by Moshi alongside the speech as predictions to compare to the ground-truth.
As reference, we also report the results of ``stage 3'' PaliGemma, as reported in \cite{beyer2024paligemma}, which starts from the same vision encoder (stage 2) used as frozen backbone in \moshivis; Note, however, that in contrast to our setting, both the vision encoder and LLM are finetuned for the downstream task in stage 3 of PaliGemma. 

First, we observe promising transfer from the image to audio modality even when learning only from speechless  samples: when trained with $p_{\text{audio}} \myeq 0\%$ and  prompted in the audio domain, the model yields 38.5\% on \ocrvqa, 49.3\% on \vqa, and a \cider~\cite{vedantam2015cider} score of 113 on \coco. 
Interestingly, the reverse is not true: at $p_{\text{audio}} \myeq 100\%$, the text evaluation performance is negatively impacted, highlighting the benefits of a mixed data supervision strategy.
However, the speech produced by a model trained with $p_{\text{audio}} \myeq 0\%$ is not coherent and also of lower audio quality when comparing to the frozen Moshi backbone, as measured by their  \mosnet~\cite{mosnet} score  in \hyperref[tab:mosnet_results]{Table \ref{tab:mosnet_results}}; for qualitative examples, see  \hyperref[app:qualitative]{Appendix \ref{app:qualitative}}.
Nevertheless, the same table also shows that adding even small amounts of audio quickly recovers audio quality.
Moreover, as shown in \hyperref[fig:text_speech_results]{Figure  \ref{fig:text_speech_results}}, increasing $p_{\text{audio}}$ 
also benefits  downstream accuracy on all tasks, reaching  scores comparable to stage 3 PaliGemma~\cite{beyer2024paligemma}, even when prompting the model in audio form.
Across these tasks, we observe that $p_{\text{audio}} \myeq 25\%$ generally yields the best trade-off between downstream performance and amount of speech training data required.

\input{resources/latex/tables/mosnet_scores_not_auto}

\myparagraph{Text-to-Audio Transfer across Tasks.} Secondly, we investigate whether the same text-to-audio transfer is observed when we have imbalanced supervision \textit{across tasks}: In other words, whether one modality overpowers the other in terms of knowledge transfer.
In this setting, we train a model such that each batch has a proportion $p_{\text{coco}}\%$ of \textit{speech} samples from \coco, and $\left(100 - p_{\text{coco}}\right)\%$ of \textit{speechless} samples from \ocrvqa, while varying the ratio of \coco to \ocrvqa samples ($p_{\text{coco}}$). We then run the reverse experiment (\ie all \ocrvqa samples are only seen in speech form, and all \coco samples are speechless). We report the audio evaluation scores (\cider and accuracy) for all models on \coco and \ocrvqa in \hyperref[tab:texttoaudiomultitask]{Table \ref{tab:texttoaudiomultitask}}: 
Increasing a dataset's training data ratio in the audio domain generally has a stronger positive effect than doing so in the text domain. This is particularly visible when evaluating in audio mode (last three rows), but also noticeable in textual evaluation (first two rows). 
In addition, this phenomenon is more salient on the specialized \ocrvqa task: With 75\% of ``speechless'' \ocrvqa training data (first column), the model  reaches an accuracy of 36.8\%. In contrast, when the same amount of \ocrvqa training data is only seen in audio form (second column), the final accuracy is 66.1\%. 

\input{resources/latex/tables/text_transfer_cider_ocrvqa_not_auto}

To verify how much audio is needed to recover performance, we then vary both the \coco to \ocrvqa ratio ($p_{\text{coco}}$) and the global ratio of audio samples ($p_{\text{audio}})$, such that we have a percentage $p_{\text{audio}} \times (100 - p_{\text{coco}})\ \%$ of spoken \ocrvqa samples in each training batch. 
As seen in \hyperref[tab:cocoocwrapup]{Table \ref{tab:cocoocwrapup}}, this immediately boosts downstream performance: 10\% of spoken \ocrvqa samples yields an accuracy of 60.7\% as compared to the previous score of  36.8\% when no audio samples were present.
In summary, the insights from the previous single-task analysis  generalizes to this two-task scenario: Mixing speech and speechless samples in every training batch is beneficial for downstream performance in both text and audio evaluation, and a ratio of $p_{\text{audio}} = 25\%$ appears to be a good trade-off between performance and amount of training speech data needed.

\input{resources/latex/tables/coco_ocr_wrapup}

\myparagraph{Ablations: Shared layers and Gating.} 
\label{sec:ablations}
While the gating mechanism described in  \hyperref[sec:train]{Section \ref{sec:train}} is  primarily introduced as a way to facilitate context switch (\hyperref[sec:context_switching]{Section  \ref{sec:context_switching}}), we first verify whether it affects the model's performance. 
To this end, we perform an ablation experiment in which we vary over \textbf{(i)} whether the adaption modules have a gating mechanism, \textbf{(ii)} whether the gate parameters are shared across layers, and  \textbf{(iii)} whether the QKV projections of cross-attention layers are shared across layers, or only the KV projections; Note that for \textbf{(iii)}, even when the projection parameters are shared, the input normalization layers to the cross-attention never are.
We report the results  in \hyperref[tab:gateablation]{Table \ref{tab:gateablation}} for \ocrvqa and in \hyperref[app:extraablation]{Appendix \ref{app:extraablation}} for \coco. 
Overall, all settings perform similarly and there is no clear winning trend across all evaluation benchmarks. 
In other words, the model is robust to design choices regarding the gate and sharing of adaptation parameters when it comes to downstream task performance alone. 
All other results reported in the paper use no parameter sharing in the gating modules, and full parameter sharing (QKV) across layers for cross-attention. In the next section, we further investigate the impact of the gating on the model's context switching abilities.

\input{resources/latex/tables/ablation_gate}

%% file: resources/latex/tables/mosnet_scores_not_auto.tex
\begin{table}[h]
    \centering
        \begin{tabular}{lcccc|c}
        \toprule
       \small \textbf{$p_{\text{audio}}$}  &\small  \textbf{0\%} &\small  \textbf{1\%} &\small  \textbf{5\%} &\small  \textbf{10\%} &\small  \textbf{Moshi} \\
        \midrule
        \small \mosnet & \small 2.78 & \small 3.59 & \small 3.47 & \small 3.56 & \small 3.34 \\
        \bottomrule
    \end{tabular}
        \caption{\textbf{Audio quality as a function of the proportion of speech samples used in training}. For each model, we evaluate its MOSNet \cite{mosnet} scores on 1000 randomly generated audio samples of roughly 40 seconds. While training with \textit{no audio} severely impacts speech quality, it quickly recovers to the same quality level as the backbone model even with just a few speech samples; for qualitative audio samples, see \hyperref[app:qualitative]{Appendix \ref{app:qualitative}}.
    \vspace{-0.25cm}
    }
    \label{tab:mosnet_results}
\end{table}

%% file: resources/latex/tables/text_transfer_cider_ocrvqa_not_auto.tex
\begin{table}[ht]
\centering
    \tikzset{bottomarrow/.style={-stealth,shorten >=100,  line width=.3mm}}
    \tikzset{toparrow/.style={-stealth,shorten <=100, line width=.3mm}}
    \resizebox{0.9\linewidth}{!}{
    \begin{tikzpicture}
        \node[] (t)
            {
\begin{tabular}{llcccccc}
\toprule
 \multicolumn{2}{l}{\textbf{$p_{\text{coco}}$ (\%)}} & \multicolumn{2}{c}{\textbf{25\%}} & \multicolumn{2}{c}{\textbf{50\%}} & \multicolumn{2}{c}{\textbf{75\%}} \\\cmidrule(lr){3-4}\cmidrule(lr){5-6}\cmidrule(lr){7-8}
 \multicolumn{2}{l}{{\raisebox{1.5ex}{\rotatebox{180}{$\Lsh$}}
given as}} & audio & text & audio & text & audio & text \\
\midrule
\multirow[c]{3}{*}{\rotatebox{90}{\textbf{text}}}\\[-.75em]
 & \coco & \cellcolor[HTML]{7cc87c} \color[HTML]{000000} 109 & \cellcolor[HTML]{00441b} \color[HTML]{f1f1f1} 121 & \cellcolor[HTML]{9bd696} \color[HTML]{000000} 107 & \cellcolor[HTML]{005221} \color[HTML]{f1f1f1} 120 & \cellcolor[HTML]{f7fcf5} \color[HTML]{000000} 98 & \cellcolor[HTML]{006027} \color[HTML]{f1f1f1} 119 \\\\[-.95em]
 & OCR & \cellcolor[HTML]{00441b} \color[HTML]{f1f1f1} 67.1 & \cellcolor[HTML]{ecf8e8} \color[HTML]{000000} 48.5 & \cellcolor[HTML]{005c25} \color[HTML]{f1f1f1} 65.6 & \cellcolor[HTML]{e1f3dc} \color[HTML]{000000} 49.8 & \cellcolor[HTML]{157f3b} \color[HTML]{f1f1f1} 63.0 & \cellcolor[HTML]{f7fcf5} \color[HTML]{000000} 46.9 \\\\[-.75em]
 \midrule
\multirow[c]{4}{*}{\rotatebox{90}{\textbf{audio}}}\\[-.75em]
 & \coco & \cellcolor[HTML]{218944} \color[HTML]{f1f1f1} 115 & \cellcolor[HTML]{f7fcf5} \color[HTML]{000000} 90 & \cellcolor[HTML]{00441b} \color[HTML]{f1f1f1} 123 & \cellcolor[HTML]{eaf7e6} \color[HTML]{000000} 93 & \cellcolor[HTML]{005723} \color[HTML]{f1f1f1} 121 & \cellcolor[HTML]{b7e2b1} \color[HTML]{000000} 100 \\\\[-.95em]
 & OCR & \cellcolor[HTML]{d4eece} \color[HTML]{000000} 36.8 & \cellcolor[HTML]{00441b} \color[HTML]{f1f1f1} 66.1 & \cellcolor[HTML]{d3eecd} \color[HTML]{000000} 37.0 & \cellcolor[HTML]{00491d} \color[HTML]{f1f1f1} 65.4 & \cellcolor[HTML]{f7fcf5} \color[HTML]{000000} 29.6 & \cellcolor[HTML]{016e2d} \color[HTML]{f1f1f1} 61.3 \\\\[-.75em]
  &\color[RGB]{100, 100, 100} MOSNet &\color[RGB]{100, 100, 100} 3.45 &\color[RGB]{100, 100, 100} 3.49 & \color[RGB]{100, 100, 100} 3.38 &\color[RGB]{100, 100, 100} 3.50 &\color[RGB]{100, 100, 100} 3.47 & \color[RGB]{100, 100, 100} 3.30 \\
\bottomrule
\end{tabular}
        };
        \draw[toparrow] (t.north west) -- (t.north east) node[midway,above] {\hspace{6.2cm} More \coco samples};
        \draw[bottomarrow] (t.south east) -- (t.south west) node[midway,below] {More \ocrvqa samples};
    \end{tikzpicture}
    } 
    \caption{\textbf{Text-to-Audio transfer with task imbalance.} We vary the audio-to-text proportion $p_{\text{coco}}$     in multi-task training, s.t. each task appears only in a single modality (\eg, \coco as audio, OCR as text; and vice-versa). We report the \cider score for  \coco and accuracy for OCR.
    We observe that task knowledge  transfers better through audio than through text. This effect is more striking when querying the model in audio at evaluation, and also more visible on the specialized task of OCR compared to COCO captioning. 
            \vspace{-0.1cm}
    }
    \label{tab:texttoaudiomultitask}
\end{table}

%% file: resources/latex/tables/coco_ocr_wrapup.tex
\begin{table}[htb]
\centering
\resizebox{\linewidth}{!}{
\bgroup
\def\arraystretch{1.2}
\begin{tabular}{llccccccccc}
\toprule 
\multicolumn{2}{l}{\textbf{$p_{\text{coco}}$(\%)}} & \multicolumn{3}{c}{\textbf{25\%}} & \multicolumn{3}{c}{\textbf{50\%}} & \multicolumn{3}{c}{\textbf{75\%}} \\
\cmidrule(lr){3-5}\cmidrule(lr){6-8}\cmidrule(lr){9-11}
\multicolumn{2}{l}{global $p_{\text{audio}}$} &  10\% & 25\% & 50\% & 10\% & 25\% & 50\% & 10\% & 25\% & 50\% \\
\midrule
\multirow[c]{2}{*}{\rotatebox{90}{\textbf{text\ \ }}} & \coco & \cellcolor[HTML]{98d594} \color[HTML]{000000} 125 & \cellcolor[HTML]{f7fcf5} \color[HTML]{000000} 123 & \cellcolor[HTML]{98d594} \color[HTML]{000000} 125 & \cellcolor[HTML]{4bb062} \color[HTML]{f1f1f1} 126 & \cellcolor[HTML]{157f3b} \color[HTML]{f1f1f1} 127 & \cellcolor[HTML]{157f3b} \color[HTML]{f1f1f1} 127 & \cellcolor[HTML]{4bb062} \color[HTML]{f1f1f1} 126 & \cellcolor[HTML]{00441b} \color[HTML]{f1f1f1} 128 & \cellcolor[HTML]{98d594} \color[HTML]{000000} 125 \\\\[-1.15em]
 & OCR & \cellcolor[HTML]{005020} \color[HTML]{f1f1f1} 68.4 & \cellcolor[HTML]{00491d} \color[HTML]{f1f1f1} 68.5 & \cellcolor[HTML]{00441b} \color[HTML]{f1f1f1} 68.6 & \cellcolor[HTML]{52b365} \color[HTML]{f1f1f1} 66.4 & \cellcolor[HTML]{5ab769} \color[HTML]{f1f1f1} 66.3 & \cellcolor[HTML]{4aaf61} \color[HTML]{f1f1f1} 66.5 & \cellcolor[HTML]{e7f6e3} \color[HTML]{000000} 63.9 & \cellcolor[HTML]{e7f6e3} \color[HTML]{000000} 63.9 & \cellcolor[HTML]{f7fcf5} \color[HTML]{000000} 63.3 \\\\[-1.15em]
 \midrule\\[-1.15em]
\multirow[c]{2}{*}{\rotatebox{90}{\textbf{audio\ }}} & \coco & \cellcolor[HTML]{f7fcf5} \color[HTML]{000000} 117 & \cellcolor[HTML]{f7fcf5} \color[HTML]{000000} 117 & \cellcolor[HTML]{73c476} \color[HTML]{000000} 120 & \cellcolor[HTML]{aedea7} \color[HTML]{000000} 119 & \cellcolor[HTML]{00441b} \color[HTML]{f1f1f1} 123 & \cellcolor[HTML]{00441b} \color[HTML]{f1f1f1} 123 & \cellcolor[HTML]{dbf1d6} \color[HTML]{000000} 118 & \cellcolor[HTML]{0b7734} \color[HTML]{f1f1f1} 122 & \cellcolor[HTML]{0b7734} \color[HTML]{f1f1f1} 122 \\\\[-1.15em]
 & OCR & \cellcolor[HTML]{6bc072} \color[HTML]{000000} 60.7 & \cellcolor[HTML]{005723} \color[HTML]{f1f1f1} 65.4 & \cellcolor[HTML]{00441b} \color[HTML]{f1f1f1} 66.1 & \cellcolor[HTML]{b2e0ac} \color[HTML]{000000} 58.4 & \cellcolor[HTML]{248c46} \color[HTML]{f1f1f1} 63.2 & \cellcolor[HTML]{0c7735} \color[HTML]{f1f1f1} 64.2 & \cellcolor[HTML]{f7fcf5} \color[HTML]{000000} 54.8 & \cellcolor[HTML]{c3e7bc} \color[HTML]{000000} 57.8 & \cellcolor[HTML]{6bc072} \color[HTML]{000000} 60.7 \\
\bottomrule
\end{tabular}
\egroup
}
    \caption{\textbf{Varying the task and speech samples proportion.} As for the single-task results (\hyperref[fig:text_speech_results]{Figure \ref{fig:text_speech_results}}), adding small amounts of audio data quickly boosts performance in both text and audio evaluation in the two-task setup: For instance, for \ocrvqa, 10\% of training audio samples yields  60.7\% accuracy, against 36.8\% when no audio samples is present for the same task ratio (\hyperref[tab:texttoaudiomultitask]{Table  \ref{tab:texttoaudiomultitask}}).
    \vspace{-0.5cm}
    }
    \label{tab:cocoocwrapup}
\end{table}

%% file: resources/latex/tables/ablation_gate.tex
\begin{table}[ht]
    \centering
\centering
  \setlength\extrarowheight{-4pt}
\resizebox{\linewidth}{!}{
\begin{tabular}{lccc@{\hskip .75cm}ccc} 
\toprule
 \ \ \ \ \ \textbf{Sharing}&\multicolumn{3}{l}{\hspace{2em}\textbf{text eval.}}&\multicolumn{3}{l}{\hspace{1.25em}\textbf{audio eval.}}
 \\
 \cmidrule(lr){2-4}\cmidrule(lr){5-7}
 {$\downarrow$ Gate / CA $\rightarrow$} & {none} & {KV} & {QKV} & {none} & {KV} & {QKV} \\
\midrule
 none & \color[HTML]{067230} 66.1 & \color[HTML]{000000}  -& \color[HTML]{000000}  - & \color[HTML]{88ce87} 63.7 & \color[HTML]{000000} - & \color[HTML]{000000} - \\
  not shared & \color[HTML]{000000} - & \color[HTML]{00441b} 67.7 & \color[HTML]{00441b} 68.2 & \color[HTML]{000000} - & \color[HTML]{03702e} 66.2 & \color[HTML]{45ad5f} 64.7 \\
  shared & \color[HTML]{000000}  -& \color[HTML]{00441b} 67.5 & \color[HTML]{087432} 66.1& \color[HTML]{000000} - & \color[HTML]{43ac5e} 64.7 & \color[HTML]{309950} 65.2 \\
\bottomrule
\end{tabular}
}\\[0.1cm]
    \caption{\textbf{Ablation on the gate and shared parameters} in the cross-attention (CA) module for \ocrvqa; for \coco results, see \hyperref[app:extraablation]{Appendix \ref{app:extraablation}}. Specifically, we evaluate different gatings (rows) and parameter sharing configurations for the CA module (columns).  Overall, there is no clear winning trend across all evaluation benchmarks: The model is robust to design choices regarding the gate and sharing of adaptation parameters when it comes to downstream task performance alone. 
        In \hyperref[sec:context_switching]{Section \ref{sec:context_switching}}, we further investigate the impact of the gate on context switching. 
    \vspace{-0.25cm}}
    \label{tab:gateablation}
\end{table}

%% file: sections/4-2-contextswitch.tex
In this section and the next, we investigate the performance of MoshiVis as a dialogue model. 
First, we quantitatively assess the model's robustness when switching between a topic relevant to the input image and a non-relevant one. In particular, we investigate how this behaviour is affected by the gating mechanism introduced in \hyperref[sec:train]{Section \ref{sec:train}}.

\myparagraph{Experimental Setup.} 
For  this section and the next (\hyperref[sec:mixmodel]{Sec. \ref{sec:mixmodel}}), we train MoshiVis as a visual dialogue model on a mix of datasets summarized in \hyperref[app:datasetoverview]{Appendix \ref{app:datasetoverview}}, including \textbf{(i)} spoken visual dialogues, \textbf{(ii)} speechless visual dialogues, and \textbf{(iii)} speechless data on specialized tasks.
For \textbf{(i)}, we generate a first set of high-quality visual dialogues for which we use human-annotated captions from the \pixmo~\cite{molmo2024} and \docci~\cite{OnoeDocci2024} datasets in the instruct prompt of the data generation pipeline described in \hyperref[sec:data]{Section \ref{sec:data}}. 
For \textbf{(ii)}, we generate similar dialogues but using captions from the PixelProse dataset~\cite{singla2024pixels}: As these captions were generated by a VLM, they tend to  contain more biases and hallucinations, hence the distinction from \pixmo and \docci. 
Finally \textbf{(iii)} is composed of publicly available benchmarks, in their original textual form, with a focus on counting and OCR tasks.

\myparagraph{Quantitative Evaluation.} 
We attempt to evaluate the robustness to context switches in a controlled, although artificial,  setting: we evaluate the model's performance on downstream tasks when presented with different   irrelevant prefixes in its context. 
More specifically, we first evaluate the ``visual to non-visual'' (\textbf{V$\rightarrow$NV}) switch by measuring the model's MMLU performance relative to that of the Moshi backbone after seeing a conversation about an image. To mimic this past conversation, we prefix the MMLU question with a random image-relevant conversation of varying length, generated with our visual dialogue data pipeline. 
Similarly, for the reverse ``non-visual to visual'' switch (\textbf{NV$\rightarrow$V}), we evaluate the model's visual performance on \coco, with a random prefixed non-image related conversation, generated in the same way as the data augmentation described at the end of \hyperref[sec:data]{Section \ref{sec:data}}.

We perform these experiments for different values of $p_{\text{concat}}$, which is the probability of prefixing/suffixing image dialogues with irrelevant conversations during training, and with different architecture choices: \textbf{(i)} no gating, \textbf{(ii)} with the gating mechanism introduced in \hyperref[sec:train]{Section \ref{sec:train}}, and \textbf{(iii)} with the gating parameters shared across all layers. 
We report the relative performance of all three  configurations in \hyperref[fig:contextswitch]{Figure \ref{fig:contextswitch}} for both the ``\textbf{V $\rightarrow$ NV}'' and the ``\textbf{NV $\rightarrow$ V}'' settings.
First, we notice that having training samples concatenated with non-image relevant prefix/suffix conversations  ($p_{\text{concat}} \mygt 0$) is   beneficial to context switch robustness, in particular when there is no gating mechanism. It sometimes even leads to improvement  when the prefix length increases, as this setting is now in-distribution respective to training. 
Similarly, the introduction of the gating mechanism  improves robustness, particularly when $p_{\text{concat}} \myeq 0$, but also interacts well with higher values of $p_{\text{concat}}$. 
Interestingly, sharing the parameters of the gate across layers sometimes even outperforms the per-layer gating model, leading to a more parameter-efficient solution.  
Finally, we provide qualitative examples of patterns learned by the gating mechanism during context switch, see \hyperref[app:qualitative]{Appendix \ref{app:qualitative}}.

\input{resources/latex/figures/contxt_switch}

%% file: resources/latex/figures/contxt_switch.tex
\begin{figure}
    \centering
    \includegraphics[width=0.98\linewidth]{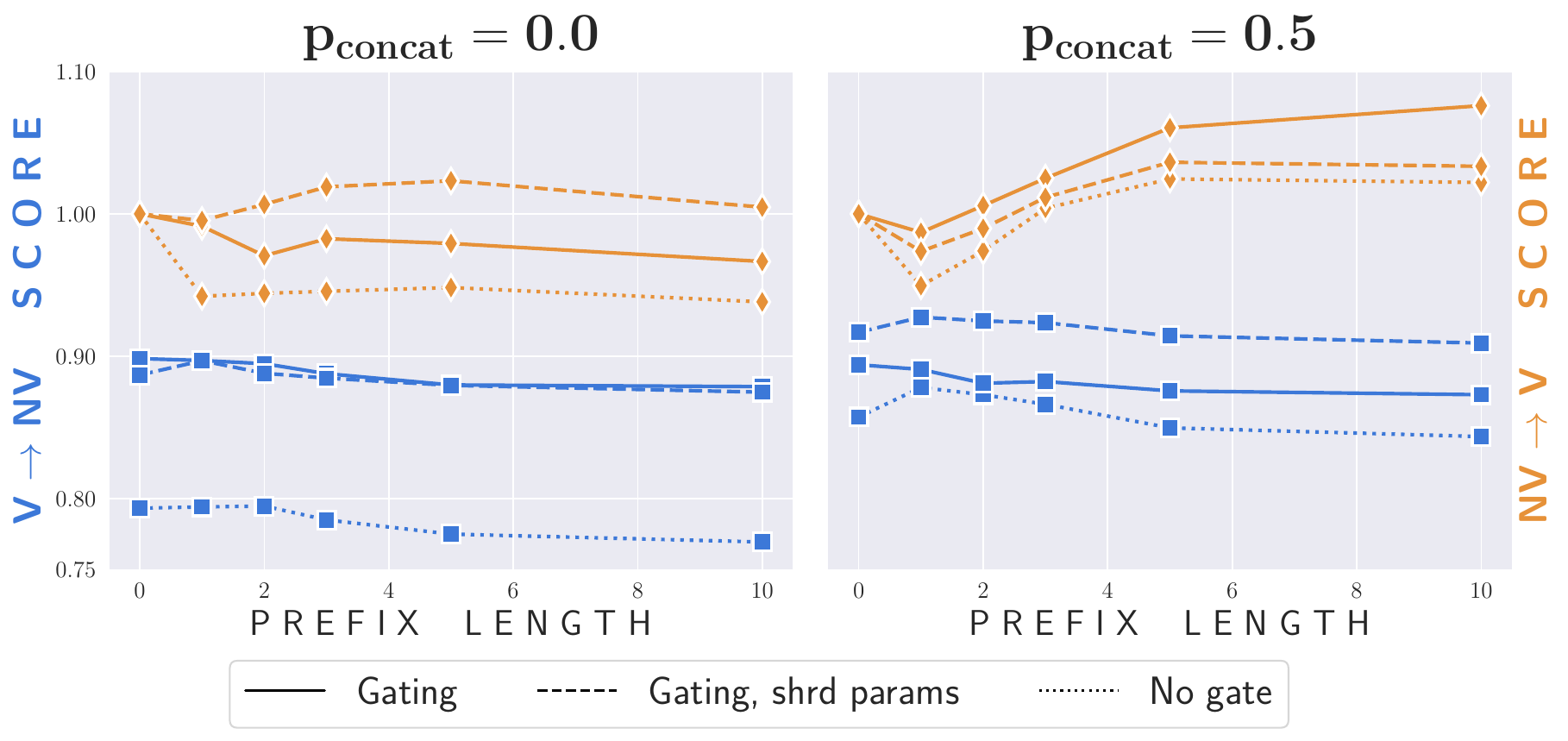}
        \caption{\textbf{Context Switch Ablation}. To assess the impact of data augmentation (\textit{left vs.~right}) as well as the gating configuration (\textit{different line styles}), we prefix every MMLU question with a randomly sampled conversation about an image ({\color[RGB]{111,148,230}V$\rightarrow$NV}), and every COCO captioning question with a randomly sampled general discussion  ({\color[RGB]{237,109,82}NV$\rightarrow$V}). We report  the model's relative performance  as a function of the random prefix length's (expressed in number of question-answer turns). We find that both data augmentation and gating improve the model's robustness to context switching.
    \vspace{-0.25cm}}
    \label{fig:contextswitch}
\end{figure}

%% file: sections/4-3-realworld.tex
We now briefly discuss the qualitative behaviour of \moshivis as a visual dialogue model.
The corresponding inference code and model weights are available at \githublink. 

\myparagraph{Latency.} 
To deploy the model, we augment the Rust and MLX backends of the open-source release of Moshi~\cite{kyutai2024moshi} with our gated adaptation modules.
On an NVIDIA L4 GPU, for images of 448 pixels (1024 tokens) and an 8-bit quantized model, MoshiVis requires  roughly 7 extra milliseconds of runtime per inference step compared to the backbone model, for a total of 51ms per step at the beginning of the conversation and 59ms with a 5-minute context window. 
We observe similar latency comparisons when testing the MLX backend on a Mac Mini with an M4 pro chip (see  \hyperref[app:latency]{Appendix \ref{app:latency}}). In both settings, the model is well within the 80ms threshold for real-time latency (the audio codec having a frequency of 12.5Hz). 
As for training time, our visual dialogue models are trained for 50k steps with batch size 64, taking roughly one day of training on 8$\times$H100 GPUs.

\myparagraph{Qualitative Results.} 
Along with this work, we provide various qualitative samples to show specific behaviours of MoshiVis. For more information, please see \hyperref[app:qualitative]{Appendix \ref{app:qualitative}}.

%% file: sections/conclusions.tex
\section{Conclusions}

Combining the three image, text and audio modalities in a unified visual speech dialogue model is a challenging problem. 
Current solutions in the open-source space are scarce and often focus on joint pre-training strategies and data selection for training such models which can be difficult to reproduce. 
In this work, we instead focus on lightweight finetuning, combining recent approaches in speech dialogue models and vision-language perceptual augmentation techniques. 
At training time, we leverage a mixture of speech and speechless (text-only) samples to learn the image-speech alignment with little audio supervision. An additional gating mechanism helps the model to switch context between visual and non-visual conversation topics.
At inference, we first evaluate the model on downstream visual performance in both text and audio form, then train it with synthetic visual dialogues that we generate, to imbue it with the ability to freely converse about both images and more general conversation topics.

\paragraph{Acknowledgements.} 
This project is funded by Iliad Group, CMA CGM Group and Schmidt Sciences. The authors thank Edouard Grave for his support and feedback throughout. They also thank Hervé Jégou for his feedback in the early phase of the project. 

%% file: sections/5-appendix.tex
\appendix

\section{Benchmark datasets}
\label{app:expsetup}

For benchmarking the visual understanding of our trained models, we use the following classical benchmarks. 

\myparagraph{Optical Character Recognition (OCR).} 
We evaluate the model's ability to recognize text in images on the
OCR-VQA~\cite{mishraICDAR19} dataset. We report the accuracy as a metric. 

\myparagraph{Visual Question Answering (VQA).} 
We evaluate the model's ability to answer general free-form questions about images on the
VQAv2~\cite{goyal2017vqav2} dataset and report the VQA accuracy as the primary metric.

\myparagraph{Image Captioning.} 
We evaluate the model's ability to generate captions for images on the COCO Captions~\cite{lin2014mscoco} dataset. 
We report the CIDEr~\cite{vedantam2015cider} score a as  metric. Specifically, we use the 2014 subset of COCO-Captions with Karpathy train/validation splits and annotations.

\section{Audio Evaluation}
\label{app:audioeval}

\subsection{Audio Benchmarks}
\label{app:ttseverything}

To query the model in audio form, we   convert the aforementioned three datasets to speech using the same text-to-speech model as in \cite{kyutai2024moshi}. We use a variety of voices for the user asking the benchmark question. 
Note that this brings a new challenge inherent to VSMs compared to VLMs, as the model's understanding of a question may vary based on the user's audio volume, intonation, accent, \etc, thus adding another level of variation compared to textual prompting. 

Note that since the frozen backbone speech model we use was initially trained as a dialogue model, we also reformat  these datasets as short conversations rather than single questions.
For instance, a simple COCO training caption such as \textit{``A boy holding an umbrella"} is converted to a spoken dialogue with the following transcript \textit{``}\texttt{[Assistant]} \textit{Hey, how are you doing?} \texttt{[User]} \textit{So, what do you see in the image?} \texttt{[Assistant]} \textit{I see a boy holding an umbrella''}.

Similarly, for the validation/test splits of benchmarks, provided on our project page, we generate speech questions which we use to query the model to perform audio  evaluation. For instance, for \coco, this can be  a dialogue of the form \textit{``}\texttt{[Assistant]} \textit{Hey, how are you?} \texttt{[User]} \textit{Can  you tell me what is in the image?''} 


\begin{table}
\begin{minipage}[c]{0.25\linewidth}
\includegraphics[width=\linewidth]{resources/image-2.png}
\end{minipage}\hfill\begin{minipage}[c]{0.7\linewidth}
\textbf{Example 1:}

\texttt{\moshivis-conversational}: ``two teddy bears in a store, one in a blue Hawaiian shirt with a brown ribbon, the other in a brown shirt with a blue ribbon''

\vspace{0.1cm}
\texttt{\moshivis-downstream}: ``Two teddy bears are on display in a store''

\end{minipage}
\vspace{.3cm}
\hrule
\vspace{0.3cm}
\begin{minipage}[c]{0.25\linewidth}
\includegraphics[width=\linewidth]{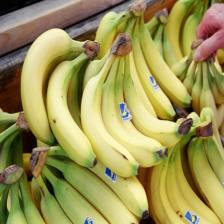}
\end{minipage}\hfill\begin{minipage}[c]{0.7\linewidth}
\textbf{Example 2:}

\texttt{\moshivis-conversational}: ``a close-up of a bunch of bananas, with a hand reaching in to pick one, and a blue sticker on one of them''

\vspace{0.1cm}
\texttt{\moshivis-downstream}: ``A bunch of bananas that are in a bin''
\end{minipage}

\vspace{.3cm}
\hrule
\vspace{0.3cm}
\begin{minipage}[c]{0.25\linewidth}
\includegraphics[width=\linewidth]{resources/image-3.png}
\end{minipage}\hfill\begin{minipage}[c]{0.7\linewidth}
\textbf{Example 3:}

\texttt{\moshivis-conversational}: ``a young boy in a baseball uniform, mid-action, with a baseball glove on his right hand''

\vspace{0.1cm}
\texttt{\moshivis-downstream}: ``A young boy in a field of grass holding a catchers mitt''
\end{minipage}
\caption{\textbf{Examples of generated COCO captions} for a conversational \moshivis (\textit{top rows}) and a \moshivis directly trained for COCO captioning as a downstream task (\textit{bottom rows}). While both models yield qualitatively accurate captions, the conversational \moshivis tends to be more verbose by nature. This can lead to lower CIDEr scores on the \coco dataset, as the score  is impacted by the length of the predicted captions.}
\label{tab:qualcoco}
\end{table}

\subsection{Formatting Challenges}
\label{app:formatting}

We observe interesting challenges during audio evaluation of \moshivis, stemming from the facts \textbf{(i)} that many text-based evaluation metrics are very sensitive to the output formatting and \textbf{(ii)} that making a model more conversational sometimes hurts its ability to be a good ``one-shot'' answerer, which is the setup of many VLM benchmarks.

For instance, \ocrvqa contains many textual signals such as punctuations for which no equivalent exists in audio; hence these may not appear in the output text stream of the model, which hurts accuracy. 
In addition, our synthetic visual dialogues are generated to give our conversational model a friendly and helpful personality, thus have a certain bias toward `yes' answers, which can be hurtful for yes/no questions present in \ocrvqa (\eg, \textit{``Is this book related to Science-Fiction?"})
As a result, for comparison, our final conversational model has an \ocrvqa accuracy of 53.3\% in audio form and 60 \% in text form, as opposed to 66.7 \% in audio form and 67.4 \% in text form when \moshivis is  directly trained for downstream performance on \ocrvqa, without seeing any conversational data. 

Similarly, CIDEr scores~\cite{vedantam2015cider} on COCO  strongly depend on the length of the generated captions. 
This often puts conversational models at a disadvantage as they tend to be more verbose and also sometimes use  ``filler'' words (\eg,`hey', `well', `so', \etc). 
For instance, our  conversational \moshivis typically reaches CIDEr scores of roughly 80 (as opposed to $\sim 125$ scores when trained on COCO) due to generating much more verbose, yet qualitatively correct, descriptions, as illustrated in \hyperref[tab:qualcoco]{Table \ref{tab:qualcoco}}. 

\section{Qualitative Samples and Behaviour}
\label{app:qualitative}
To further support the findings discussed in this work, we provide additional qualitative samples on our project page at \githublink.

First, as discussed in \hyperref[sec:text_transfer]{Section \ref{sec:text_transfer}}, we observe that the MOSNet scores for measuring audio quality of the model strongly improves when adding even a small amount of audio samples during training. As we show on the project page (in the section ``Impact of Speechless Data on Audio Quality"), this can also be observed qualitatively on the generated speech samples.

{Moreover, we provide various qualitative samples of real conversations with MoshiVis trained as a visual dialogue system, in order highlight specific behaviours of the model. This includes, \eg, the general ability to hold visual conversations, specific skills such as reading and counting, and the ability to switch contexts or speak in different voices.

Lastly, to better understand the gating learned during training, we also provide samples for which we visualize the aggregated (averaged across layers) per-token gating values used by the model; see also \hyperref[fig:furby]{Figure \ref{fig:furby}} for an example.

\section{Additional results}

\subsection{Gate Ablation}
\label{app:extraablation}

In \hyperref[tab:gateablationfull]{Table \ref{tab:gateablationfull}}, similar to \hyperref[tab:gateablation]{Table \ref{tab:gateablation}}, we report results on \coco for different configurations for the gating and sharing of parameters in the cross-attention modules. We find that the insights observed on the \ocrvqa dataset also apply to the \coco experiments. Specifically, the model's benchmark performance is robust to these design choices and there is no clear ``winning configuration''.
\input{resources/latex/tables/full_ablation_gate_full}

\subsection{Latency with MLX backends}
\label{app:latency}
In \hyperref[fig:mlxplots]{Figure \ref{fig:mlxplots}}, we report latency results for the MLX backend running locally on a Mac Mini with an Apple M4 pro chip.
We evaluate these latencies with our model as well as the original Moshi backbone, quantized to 8 bits with a block size of 64.

\begin{figure}[!htb]
    \centering
    \includegraphics[width=0.9\linewidth]{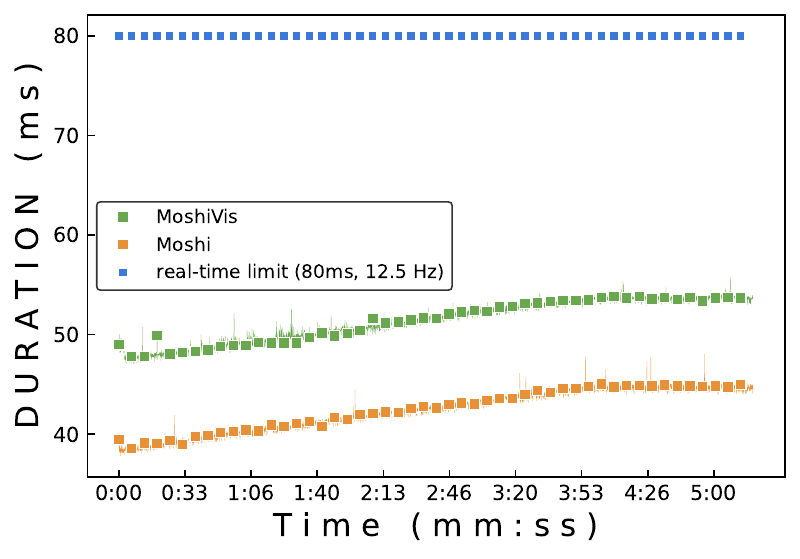}
    \caption{\textbf{Latency results with the MLX backend} on a Mac Mini with a M4 Pro chip. Here we report the latency per inference step (time to generate one speech token) for \moshivis and the original Moshi backbone, both quantized in 8 bits. Both models stay well below the real-time limit of 80ms (12.5Hz audio codec) during a 5-minutes conversation span.
    \vspace{-0.25cm}
    }
    \label{fig:mlxplots}
\end{figure}

\section{Synthetic Data Generation Pipeline}

\subsection{Overview}
\label{app:promptengineering}

To generate the synthetic visual dialogues, we use two separate instances of Mistral's Nemo models \cite{mistralnemo}, each with its own set of instructions (`User' and `Assistant'): The user always asks questions and the assistant always answers them.

We generate a set of user-assistant instruction pairs (provided through \cref{instruct:lead,instruct:loc,instruct:nfc,instruct:prop,instruct:start,instruct:tbs,instruct:tns,instruct:num}), each characterising a specific behaviour or interaction. The instructions have been designed to endow the model with certain behavioural patterns, such as being robust to misleading questions (\cref{instruct:lead,instruct:nfc}), or to promote learning to extract certain facts from the image embeddings such as spatial information (\cref{instruct:loc}), recognising object attributes (\cref{instruct:prop}), counting (\cref{instruct:num}), or to produce general question-answer conversations (\cref{instruct:tns,instruct:tbs}).
Finally \cref{instruct:start} is a special instruction to generate the start of a generic visual dialogue (\eg, ``\textit{what's in the image?}'' in many varied ways).

\myparagraph{Instruction Template.} For each instruction,  we provide the `Instruction Template' (see, \eg, \hyperref[instruct:start]{Instruction \ref{instruct:start}}). It  is used to generate a model-specific instruction (by replacing the \texttt{\{ROLE\_SPECIFIC\_TEXT\}} with the respective texts and \texttt{\{caption\}} with the image caption). These are then provided as `system prompts' (\ie, in between \texttt{[SYS]} tags) to the Mistral Nemo models. We then force the start of the conversation by `Forced start of the conversation', which triggers the first turn of the `User' model---after that, the forced start is removed from the conversation history and the models `talk between themselves'. 

\myparagraph{Generating dialogues.}
In practice, to generate a dialogue, we can stick to a single  type of instruction throughout the whole conversation (\eg, for \cref{instruct:lead,instruct:loc,instruct:nfc,instruct:prop,instruct:tbs,instruct:tns}) for multiple turns of conversation.

Alternatively, we also generate a more generic style of conversation that combines multiple instructions, see \eg 
\hyperref[tab:dataset_overview]{Table \ref{tab:dataset_overview}}. 
For this, we first start with the instruction given in 
\hyperref[instruct:start]{Instruction \ref{instruct:start}}, which samples a generic question about the image (\eg,  \textit{``what's in the image?''}).
After  the first turn of the conversation, we then randomly sample the model instruction in each subsequent turn (question-answer pair) for the continuation of the conversation. Hence, for every conversation, the models are provided with the full history of the past conversation (excluding `forced start', and exchanging the `system prompts' for the randomly sampled ones) and prompted to continue the conversation. 

This results in combined conversations that always start with high-level description questions (\texttt{COMB}, \hyperref[instruct:start]{Instruction \ref{instruct:start}}) then ask multiple questions about various aspects of the image, \eg, location of objects (\texttt{LOC}, \hyperref[instruct:loc]{Instruction \ref{instruct:loc}}), their colors and properties (\texttt{PROP}, \hyperref[instruct:prop]{Instruction \ref{instruct:prop}}), their numbers (\texttt{NUM}, \hyperref[instruct:num]{Instruction \ref{instruct:num}}), including  misleading questions (\texttt{LEAD1}, \hyperref[instruct:lead]{Instruction \ref{instruct:lead}}; \texttt{LEAD2}, \hyperref[instruct:nfc]{Instruction \ref{instruct:nfc}}), or generic interactions between a `Teacher' and a `Student' (\texttt{TS1}, \hyperref[instruct:tns]{Instruction \ref{instruct:tns}}; \texttt{TS2}, \hyperref[instruct:tbs]{Instruction \ref{instruct:tbs}}). 

\definecolor{customgray}{RGB}{250, 250, 250}

\subsection{Final Datasets Overview}
\label{app:datasetoverview}

In \hyperref[tab:dataset_overview]{Table \ref{tab:dataset_overview}}, we describe the final datasets we use for training the conversational \moshivis. 
We sample each batch such that the relative proportion of each dataset follows the distribution given by the relative weight $\omega_i$ (third column). 

The final dataset mixture is split into three categories:
\begin{itemize}
\item First, we generate a set of high-quality visual dialogues for which we use human-annotated captions from the \pixmo~\cite{molmo2024} and \docci~\cite{OnoeDocci2024} datasets in the instruct prompt of the data generation pipeline described in \hyperref[sec:data]{Section \ref{sec:data}}: These are \texttt{\docci PROP, \docci LOC, \pixmo LEAD1, \pixmo COMB}; for the detailed instructions for PROP, LOC, LEAD1 and COMB, see \hyperref[dump:instruct]{Appendix \ref{dump:instruct}}.

\item We generate similar dialogues but using captions from the PixelProse dataset~\cite{singla2024pixels}: As these captions were generated by a VLM, they tend to  contain more biases and hallucinations, hence the distinction from \pixmo and \docci. These are \texttt{PixelProse TS1, PixelProse TS2, PixelProse LEAD2}; for the detailed instructions for TS1, TS2 and LEAD2, see \hyperref[dump:instruct]{Appendix \ref{dump:instruct}}.
\item Finally, we add non-dialogue style datasets in textual form to leverage publicly available image-text benchmarks,  with a focus on counting and OCR tasks: \texttt{TallyQA, \ocrvqa, Rendered Text, DocVQA}.
\end{itemize}
\input{resources/latex/tables/dataset_overview}

\onecolumn

\begin{figure*}[t]
\centering
\includegraphics[width=0.9\linewidth]{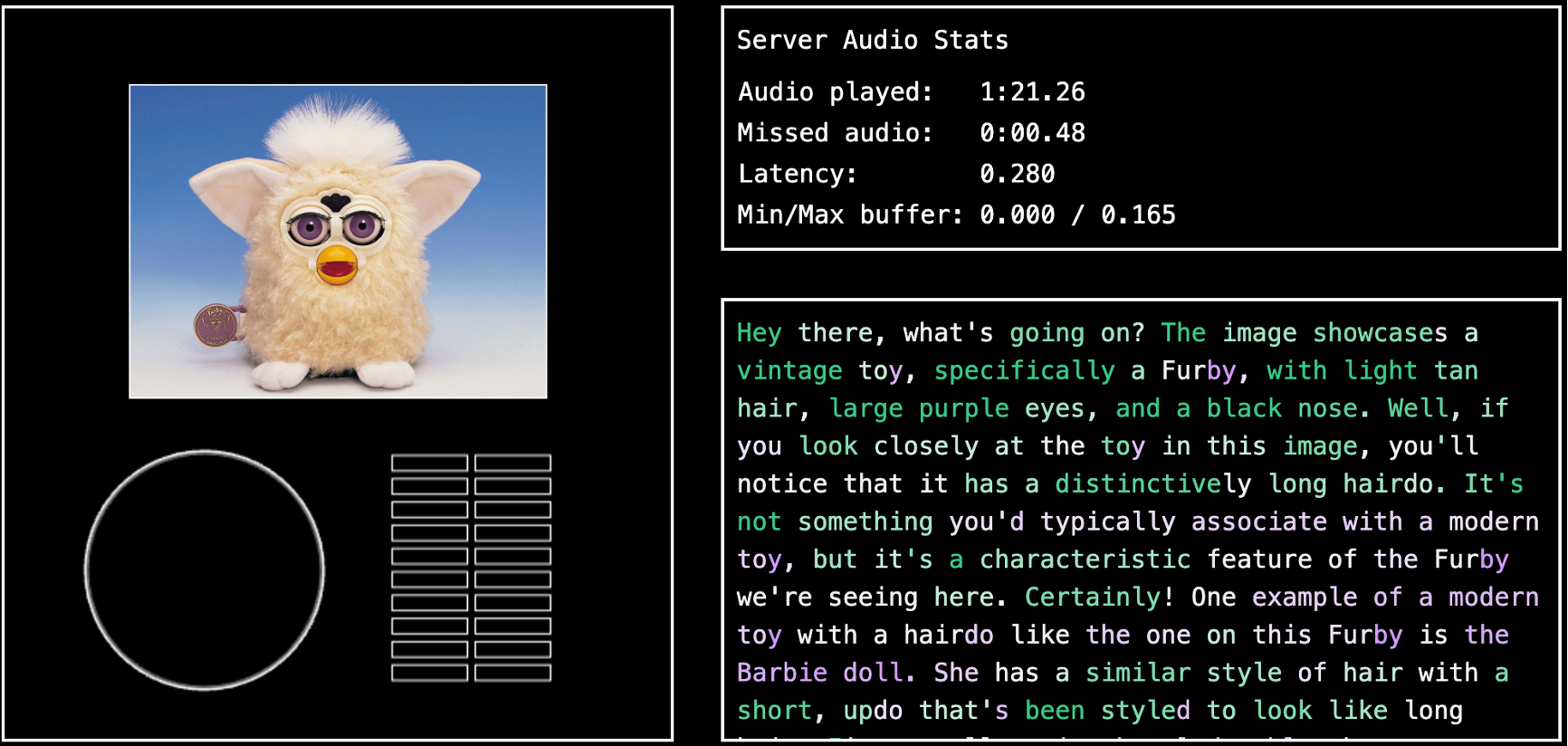}
\caption{\textbf{Example visualization of the gate activations} during a conversation about a given input image (\textit{left}). On the right, we see the text stream output by \moshivis alongside the audio tokens, which only contains the assistant's produced text tokens. We color the tokens based on the average output values of the gate sigmoid activation across all layers (\textcolor{SeaGreen}{high values} in green and \textcolor{violet}{lower values} in purple) during the conversation. We observe that, despite the absence of explicit supervision, the gate learns relevant patterns: It tends to activate more on image-relevant information, and less on more general knowledge questions. 
\label{fig:furby}}
\end{figure*}

\clearpage
\section{Detailed Instructions for Conversation Generation}
\label{dump:instruct}
In the following, we provide the detailed instructions used in our data generation pipeline, see also \hyperref[app:promptengineering]{Appendix \ref{app:promptengineering}}.

\input{resources/latex/prompts/latex_output_SSG}
\input{resources/latex/prompts/latex_output_LOC}
\input{resources/latex/prompts/latex_output_PROP}
\input{resources/latex/prompts/latex_output_NUM}
\input{resources/latex/prompts/latex_output_TBS}
\input{resources/latex/prompts/latex_output_TNS}
\input{resources/latex/prompts/latex_output_LEAD}
\input{resources/latex/prompts/latex_output_NFC}
\twocolumn

%% file: resources/latex/tables/full_ablation_gate_full.tex
\begin{table}[thb]
    \centering
\begin{minipage}[c]{0.95\linewidth}
\centering
\resizebox{\linewidth}{!}{
\begin{tabular}{lccc@{\hskip .75cm}ccc} 
\toprule
 \ \ \ \ \ \textbf{Sharing}&\multicolumn{3}{l}{\hspace{2em}\textbf{text eval.}}&\multicolumn{3}{l}{\hspace{1.25em}\textbf{audio eval.}}
 \\
 \cmidrule(lr){2-4}\cmidrule(lr){5-7}
 {$\downarrow$ Gate / CA $\rightarrow$} & {none} & {KV} & {QKV} & {none} & {KV} & {QKV} \\
\midrule
 none & \color[HTML]{006729} 126 & \color[HTML]{000000}  - & \color[HTML]{000000}  -  & \color[HTML]{016e2d} 125 & \color[HTML]{000000}  - & \color[HTML]{000000}  -\\
  not shared & \color[HTML]{000000}  - & \color[HTML]{005321} 127 & \color[HTML]{005622} 126 & \color[HTML]{000000}  - & \color[HTML]{218944} 123 & \color[HTML]{17813d} 124\\
  shared & \color[HTML]{000000}  - & \color[HTML]{006529} 126 & \color[HTML]{19833e} 124 & \color[HTML]{000000}  - & \color[HTML]{0c7735} 124 & \color[HTML]{2f974e} 122\\
\bottomrule
\end{tabular}
}\\[0.1cm]
\end{minipage}
    \caption{\textbf{Ablation on the gate and shared parameters on COCO}. 
    We report \cider scores for different configurations of the gate and the cross-attention (CA) module. As for \ocrvqa (\hyperref[sec:ablations]{Section \ref{sec:ablations}}), there is no clear winning trend across all evaluation benchmarks: The model is robust to design choices regarding the gate and sharing of adaptation parameters when it comes to downstream task performance alone.  }
    \label{tab:gateablationfull}
\end{table}

%% file: resources/latex/tables/dataset_overview.tex
\begin{table}[h]
    \centering
    \resizebox{\linewidth}{!}{
    \begin{tabular}{llcc}
        \toprule
        \textbf{Dataset Name} &\textbf{Source Dataset} & \textbf{Rel. Weight} & \textbf{Type} \\
        \midrule
                \docci PROP & \docci~\cite{OnoeDocci2024} & 5 & Speech\\
        \docci LOC & \docci~\cite{OnoeDocci2024} & 5 & Speech\\
        \pixmo LEAD1 & \pixmo~\cite{molmo2024} & 5 & Speech\\
        \pixmo COMB & \pixmo~\cite{molmo2024} & 15 & Speech\\
        \midrule
                \pixelprose TS1 & \pixelprose~\cite{singla2024pixels} & 15 & Text\\
        \pixelprose TS2 & \pixelprose~\cite{singla2024pixels} & 15 & Text\\
        \pixelprose LEAD2 & \pixelprose~\cite{singla2024pixels} & 15 & Text\\
        \midrule
        \tallyqa & \tallyqa~\cite{acharya2019tallyqa} & 1 & Text\\
        \ocrvqa & \ocrvqa~\cite{mishra2019ocr} & 5 & Text\\
        RENDERED TEXT & RenderedText~\cite{renderedtext} & 1 & Text\\
        \docvqa & \docvqa~\cite{tito2021document} & 2 & Text\\
        \bottomrule
    \end{tabular}
    }
    \caption{\textbf{Datasets used for training MoshiVis.} We list the combination of datasets we used along with the respective source datasets used to create them, the relative frequency with which we sampled them, and whether the dataset contained audio (`Type'). In particular, the datasets were sampled with a probability given by $p_\text{sample} \myeq w_i/\sum_j w_j$, with $w_i$ the relative weight (`Rel. Weight') of each split. The respective splits  were created according to the generation scripts and prompts discussed in \hyperref[app:promptengineering]{Appendix \ref{app:promptengineering}}.}
    \label{tab:dataset_overview}
\end{table}

%% file: resources/latex/prompts/latex_output_SSG.tex
\begin{center}
\begin{tcolorbox}[colback=customgray, colframe=black!50, sharp corners, boxrule=0.5pt, width=0.95\textwidth]
\subsection*{Default Starting Instructions}
\textbf{Instruction Template}
\begin{lstlisting}[breakautoindent=false,breakindent=0ex]
You take part in a casual discussion about an image. 
{ROLE_SPECIFIC_TEXT}
\end{lstlisting}
\textbf{Role-speficic text (User):}
\begin{lstlisting}
You want to learn more about the image you and the other speaker are looking at.  Your aim is to obtain a description of the image.
\end{lstlisting}
\textbf{Role-speficic text (Assistant):}
\begin{lstlisting}
The image is described in detail by the following description:
{caption}

You are a friendly and factual conversational assistant. Your task is to give a SHORT SUMMARY what you see in the image in A FEW sentences . You NEVER SAY HELLO NOR HI
\end{lstlisting}
\medskip
\textbf{Forced start of the conversation:}
\begin{lstlisting}
Start the conversation by ASKING A SINGLE question about what can be seen in the IMAGE. You use DIVERSE YET REALISTIC ways to ask your question; 

# randomly vary over question length
if (p := random.random()) < 0.5:
    "VERY IMPORTANT: your question should be LESS THAN 8 words"
elif p < 0.75:
    "VERY IMPORTANT: your question should be LESS THAN 14 words"
else:
    "VERY IMPORTANT: your question should be LESS THAN 26 words"
# radomly vary across tone
if random.random() < 0.5:
    "You ask the question in a direct style; For instance: 'What do YOU see in the image ?'\n "
else:
     "You ask the question from your own point of view; For instance: 'What am I looking at ?'\n "

if random.random() < 0.75:
    "You speak in a confident assertive tone.\n "
else:
    "You speak in a hesitant, hard to follow, manner.\n "

# Vary point of view
if random.random() < 0.5:
    "You ask what the user SEE in the image.\n "
else:
    "You ask what's visible in the image\n "
!ALWAYS ASK A SINGLE QUESION!
\end{lstlisting}
\end{tcolorbox}
\tboxcaption{\texttt{\textbf{COMB}}.
The default starting instructions themselves are used only to obtain a single turn conversation (user + assistant).
Specifically, they are designed to obtain diverse starting points for the synthetic dialogues and, in practice,
    they are COMBined with other randomly sampled instructions from \cref{instruct:lead,instruct:loc,instruct:prop,instruct:tbs,instruct:tns,instruct:nfc}
    to form a multiturn conversation.
    Note that the \texttt{if random.random()<0.5} instructions are not part of the prompt,
    but actual sampling operations are executed every time to generate the initial prompt for each  new dialogue.
}
\label{instruct:start}
\end{center}

%% file: resources/latex/prompts/latex_output_LOC.tex
\begin{center}
\begin{tcolorbox}[colback=customgray, colframe=black!50, sharp corners, boxrule=0.5pt, width=0.95\textwidth]
\subsection*{Instructions for conversations about spatial information}
\textbf{Instruction Template}
\begin{lstlisting}[breakautoindent=false,breakindent=0ex]
Image description:
"""{caption}"""
{ROLE_SPECIFIC_TEXT}
\end{lstlisting}
\textbf{Role-speficic text (User):}
\begin{lstlisting}
You are engaging in a conversation about an image with another person.
Your goal is to ask detailed questions about everything that is visible in the image, starting from the most salient features (main objects and their relationships) to finer details (the overall setting, background features, time of day, season, etc).
To guide your questions, you have been secretly provided with a detailed description of the image (see above); this fact should not be revealed however!
You will use this secret description to only ask questions that can be answered based on this description.
YOU SHOULD AVOID EASY YES/NO QUESTIONS!You do not ask leading questions that already contain or give a hint at the answer; i.e., avoid ending your question in 'isn't it'/'does it'/'doesn't it' etc.
In your questions, you emphasize the spatial relations / locations of what is in the image. You only ask about spatial relations explicitly known from the image description. If possible, ask spatial questions about different aspects of the image.

\end{lstlisting}
\textbf{Role-speficic text (Assistant):}
\begin{lstlisting}
You are a helpful conversation partner who can see the image above and is willing to describe it to another person.
You provide detailed (but not too verbose!) answers about the image in response to their questions.
When answering:
- Be detailed and factual, use simple language and keep the answer short. No matter what the other speaker is implying, you always base your answer on the true facts given in the image description.
- Be assertive about facts that are provided in the original description.
- Contradict the other speaker when adequate such as receiving information that contradicts the description.
- Speak naturally, as though you are sharing your genuine observations with someone looking at the image alongside you.
- Avoid any indication that you are relying on a description or external data. Do not use phrases like "I was told" or "Based on what I read."
- Engage in a dynamic conversation-answer questions about the image, offer additional observations, and encourage exploration of its details.
- Make thoughtful, plausible inferences when necessary, but always stay grounded in what is realistically observable in the image.
- For example, if asked about the mood of the image, consider elements like lighting, colors, facial expressions, or the setting to infer emotions.
- If asked about a specific detail, respond as if you are focusing on that part of the image directly.
- MOST IMPORTANTLY: You never invent any new facts!Your goal is to create an immersive and conversational experience, simulating the act of perceiving the image firsthand.
Remember to NEVER make up any facts about the image, answer solely based on the description provided.
\end{lstlisting}
\medskip
\textbf{Forced start of the conversation:}
\begin{lstlisting}
Start the conversation by asking a question about the image in any way you want!

\end{lstlisting}
\end{tcolorbox}
\tboxcaption{\texttt{\textbf{LOC}}. To improve factuality and better extract \textit{spatial} information from the image embeddings, we instruct the models to specifically ask questions about locations of objects and answer based only on the captions. We additionally use \cref{instruct:prop}, to extract attribute information.}
\label{instruct:loc}
\end{center}

%% file: resources/latex/prompts/latex_output_PROP.tex
\begin{center}
\begin{tcolorbox}[colback=customgray, colframe=black!50, sharp corners, boxrule=0.5pt, width=0.95\textwidth]
\subsection*{Instructions for conversations about object property information}
\textbf{Instruction Template}
\begin{lstlisting}[breakautoindent=false,breakindent=0ex]
Image description:
"""{caption}"""
{ROLE_SPECIFIC_TEXT}
\end{lstlisting}
\textbf{Role-speficic text (User):}
\begin{lstlisting}
You are engaging in a conversation about an image with another person.
Your goal is to ask detailed questions about everything that is visible in the image, starting from the most salient features (main objects and their relationships) to finer details (the overall setting, background features, time of day, season, etc).
To guide your questions, you have been secretly provided with a detailed description of the image (see above); this fact should not be revealed however!
You will use this secret description to only ask questions that can be answered based on this description.
YOU SHOULD AVOID EASY YES/NO QUESTIONS!You do not ask leading questions that already contain or give a hint at the answer; i.e., avoid ending your question in 'isn't it'/'does it'/'doesn't it' etc.
In your questions, you focus on attributes of what is visible in the image (as given via descriptions and adjectives in the image description). This includes in particular the COLOR of object, their SHAPE or their TEXTURE.  You only ask about properties explicitly known from the image description. If possible, ask  questions about different aspects of the image.

\end{lstlisting}
\textbf{Role-speficic text (Assistant):}
\begin{lstlisting}
You are a helpful conversation partner who can see the image above and is willing to describe it to another person.
You provide detailed (but not too verbose!) answers about the image in response to their questions.
When answering:
- Be detailed and factual, use simple language and keep the answer short. No matter what the other speaker is implying, you always base your answer on the true facts given in the image description.
- Be assertive about facts that are provided in the original description.
- Contradict the other speaker when adequate such as receiving information that contradicts the description.
- Speak naturally, as though you are sharing your genuine observations with someone looking at the image alongside you.
- Avoid any indication that you are relying on a description or external data. Do not use phrases like "I was told" or "Based on what I read."
- Engage in a dynamic conversation-answer questions about the image, offer additional observations, and encourage exploration of its details.
- Make thoughtful, plausible inferences when necessary, but always stay grounded in what is realistically observable in the image.
- For example, if asked about the mood of the image, consider elements like lighting, colors, facial expressions, or the setting to infer emotions.
- If asked about a specific detail, respond as if you are focusing on that part of the image directly.
- MOST IMPORTANTLY: You never invent any new facts!Your goal is to create an immersive and conversational experience, simulating the act of perceiving the image firsthand.
Remember to NEVER make up any facts about the image, answer solely based on the description provided.
\end{lstlisting}
\medskip
\textbf{Forced start of the conversation:}
\begin{lstlisting}
Start the conversation by asking a question about the image in any way you want!

\end{lstlisting}
\end{tcolorbox}
\tboxcaption{\texttt{\textbf{PROP}}.  Similar to \cref{instruct:loc}, to improve factuality and better extract \textit{attribute} information (\eg colours, textures, shapes) from the image embeddings, we instruct the models to specifically ask questions about such attributes of objects and to answer based only on the captions.}
\label{instruct:prop}
\end{center}

%% file: resources/latex/prompts/latex_output_NUM.tex
\begin{center}
\begin{tcolorbox}[colback=customgray, colframe=black!50, sharp corners, boxrule=0.5pt, width=0.95\textwidth]
\subsection*{Instructions for conversations about spatial information}
\textbf{Instruction Template}
\begin{lstlisting}[breakautoindent=false,breakindent=0ex]
Image description:
"""{caption}"""
{ROLE_SPECIFIC_TEXT}
\end{lstlisting}
\textbf{Role-speficic text (User):}
\begin{lstlisting}
You are engaging in a conversation about an image with another person.
Your goal is to ask detailed questions about everything that is visible in the image, starting from the most salient features (main objects and their relationships) to finer details (the overall setting, background features, time of day, season, etc).
To guide your questions, you have been secretly provided with a detailed description of the image (see above); this fact should not be revealed however!
You will use this secret description to only ask questions that can be answered based on this description.
YOU SHOULD AVOID EASY YES/NO QUESTIONS!You do not ask leading questions that already contain or give a hint at the answer; i.e., avoid ending your question in 'isn't it'/'does it'/'doesn't it' etc.
Your questions focus on the NUMBER of objects visible in the image. If possible, ask spatial questions about different objects categories in the image"

\end{lstlisting}
\textbf{Role-speficic text (Assistant):}
\begin{lstlisting}
You are a helpful conversation partner who can see the image above and is willing to describe it to another person.
You provide detailed (but not too verbose!) answers about the image in response to their questions.
When answering:
- Be detailed and factual, use simple language and keep the answer short. No matter what the other speaker is implying, you always base your answer on the true facts given in the image description.
- Be assertive about facts that are provided in the original description.
- Contradict the other speaker when adequate such as receiving information that contradicts the description.
- Speak naturally, as though you are sharing your genuine observations with someone looking at the image alongside you.
- Avoid any indication that you are relying on a description or external data. Do not use phrases like "I was told" or "Based on what I read."
- Engage in a dynamic conversation-answer questions about the image, offer additional observations, and encourage exploration of its details.
- Make thoughtful, plausible inferences when necessary, but always stay grounded in what is realistically observable in the image.
- For example, if asked about the mood of the image, consider elements like lighting, colors, facial expressions, or the setting to infer emotions.
- If asked about a specific detail, respond as if you are focusing on that part of the image directly.
- MOST IMPORTANTLY: You never invent any new facts!Your goal is to create an immersive and conversational experience, simulating the act of perceiving the image firsthand.
Remember to NEVER make up any facts about the image, answer solely based on the description provided.
\end{lstlisting}
\medskip
\textbf{Forced start of the conversation:}
\begin{lstlisting}
Start the conversation by asking a question about the image in any way you want!

\end{lstlisting}
\end{tcolorbox}
\tboxcaption{\texttt{\textbf{NUM}}. To improve factuality in particular about the number of objects, we put a specific emphasis on these types of questions through this instruct. We additionally use \cref{instruct:prop}, to extract attribute information and \cref{instruct:loc} for object location}
\label{instruct:num}
\end{center}

%% file: resources/latex/prompts/latex_output_TBS.tex
\begin{center}
\begin{tcolorbox}[colback=customgray, colframe=black!50, sharp corners, boxrule=0.5pt, width=0.95\textwidth]
\subsection*{Teacher-student instructions \#1}
\textbf{Instruction Template}
\begin{lstlisting}[breakautoindent=false,breakindent=0ex]
IMAGE DESCRIPTION START
{caption}
IMAGE DESCRIPTION END
You are an *external observer* having a casual dialogue about the image described above.
You pretend that you see the image itself, **under no circumstances** mention that you got the information from a description!!
{ROLE_SPECIFIC_TEXT}
You sound confident and assertive and most importantly, you always stick to the facts described!! 
Again, DO NOT ADD FACTS, DO NOT MENTION THE DESCRIPTION, DO NOT MENTION THE OTHER SPEAKER's NAME.
\end{lstlisting}
\textbf{Role-speficic text (User):}
\begin{lstlisting}
You are the student!! YOU DO NOT HAVE ACCESS TO THE DESCRIPTION so you have to get all the information from your teacher.  Your goal is to learn about everything about the image. You should refer to the image in your questions. e.g. 'is ... visible in the image' or 'Do you see ... in the image' or 'What is in the image?' You sometimes ask questions about something NOT VISIBLE IN THE IMAGE. In particular, you want to learn about the NUMBER of objects, their LOCATION and their COLOR. You ask ONLY ONE QUESTION AT A TIME!
\end{lstlisting}
\textbf{Role-speficic text (Assistant):}
\begin{lstlisting}
You are the strict teacher!! Your anwers should be complete and detailed, but NOT TOO LONG. Do not EVER mention the description. You are nice but firm and DO NOT HESITATE TO CORRECT THE STUDENT.  You never mention any facts that are not explicitly described about the image!!! NEVER mention the athmosphere of the image, only its CONTENT
\end{lstlisting}
\medskip
\textbf{Forced start of the conversation:}
\begin{lstlisting}
Start the conversation by asking a question  about an object which is NOT mentioned in the description. 
\end{lstlisting}
\end{tcolorbox}
\tboxcaption{\texttt{\textbf{TS1}}. To improve the model's robustness to all kinds of general questions about images, we designed two different sets of instructions for `student--teacher' interactions (see also \cref{instruct:tns}). Specifically, in this instruction, we instruct the student to try to learn as much as possible about the image by asking the teacher, with a particular focus on factual elements.}
\label{instruct:tbs}
\end{center}

%% file: resources/latex/prompts/latex_output_TNS.tex
\begin{center}
\begin{tcolorbox}[colback=customgray, colframe=black!50, sharp corners, boxrule=0.5pt, width=0.95\textwidth]
\subsection*{Teacher-student instructions \#2}
\textbf{Instruction Template}
\begin{lstlisting}[breakautoindent=false,breakindent=0ex]
IMAGE DESCRIPTION START
{caption}
IMAGE DESCRIPTION END
You are an *external observer* having a casual dialogue about the image described above.
You pretend that you see the image itself, **under no circumstances** mention that you got the information from a description!!
{ROLE_SPECIFIC_TEXT}
You sound confident and assertive and most importantly, you always stick to the facts described!! 
Again, DO NOT ADD FACTS, DO NOT MENTION THE DESCRIPTION, DO NOT MENTION THE OTHER SPEAKER's NAME.
\end{lstlisting}
\textbf{Role-speficic text (User):}
\begin{lstlisting}
You are the student!! You do not see the image very well and your goal is to ask simple (almost stupid) questions about the image to learn more about its content. You should refer to the image in your questions. e.g. 'is ... visible in the image' or 'Do you see ... in the image' or 'What is in the image?' Your questions should also details about the LOCATION of objects and a bit about their COLOR. You ask ONLY ONE QUESTION AT A TIME!
\end{lstlisting}
\textbf{Role-speficic text (Assistant):}
\begin{lstlisting}
You are the teacher!! Your anwers should be complete and detailed, and long. Do not EVER mention the description. You never mention any facts that are not explicitly described about the image!!! NEVER mention the athmosphere of the image, only its CONTENT
\end{lstlisting}
\medskip
\textbf{Forced start of the conversation:}
\begin{lstlisting}
Start the conversation by asking a question  about an object which is NOT mentioned in the description. 
\end{lstlisting}
\end{tcolorbox}
\tboxcaption{\texttt{\textbf{TS2}}. To improve the model's robustness to all kinds of general questions about images, we designed two different sets of instructions for `student--teacher' interactions (see also \cref{instruct:tbs}). Specifically, in this instruction, we instruct the student to ask simple (`almost stupid') questions about the image, with a particular focus on factual elements.}
\label{instruct:tns}
\end{center}

%% file: resources/latex/prompts/latex_output_LEAD.tex
\begin{center}
\begin{tcolorbox}[colback=customgray, colframe=black!50, sharp corners, boxrule=0.5pt, width=0.95\textwidth]
\subsection*{Instructions to ask misleading questions \#1}
\textbf{Instruction Template}
\begin{lstlisting}[breakautoindent=false,breakindent=0ex]
Image description:
"""{caption}"""
{ROLE_SPECIFIC_TEXT}
\end{lstlisting}
\textbf{Role-speficic text (User):}
\begin{lstlisting}
You are engaging in a conversation about an image with another person.
Your goal is to ask detailed questions about everything that is visible in the image, starting from the most salient features (main objects and their relationships) to finer details (the overall setting, background features, time of day, season, etc).
To guide your questions, you have been secretly provided with a detailed description of the image (see above); this fact should not be revealed however!
You will use this secret description to only ask questions that can be answered based on this description.
YOU SHOULD AVOID EASY YES/NO QUESTIONS!You do not ask leading questions that already contain or give a hint at the answer; i.e., avoid ending your question in 'isn't it'/'does it'/'doesn't it' etc.
In your questions, you often BUT NOT ALWAYS try to mislead the other speaker into believing something that is not correct.
For instance, you ask about a RANDOM object not in the image but keep your questions short!! You should be almost rude in your questions. 
\end{lstlisting}
\textbf{Role-speficic text (Assistant):}
\begin{lstlisting}
You are a helpful conversation partner who can see the image above and is willing to describe it to another person.
You provide detailed (but not too verbose!) answers about the image in response to their questions.
When answering:
- Be detailed and factual, use simple language and keep the answer short. No matter what the other speaker is implying, you always base your answer on the true facts given in the image description.
- Be assertive about facts that are provided in the original description.
- Contradict the other speaker when adequate such as receiving information that contradicts the description.
- Speak naturally, as though you are sharing your genuine observations with someone looking at the image alongside you.
- Avoid any indication that you are relying on a description or external data. Do not use phrases like "I was told" or "Based on what I read."
- Engage in a dynamic conversation-answer questions about the image, offer additional observations, and encourage exploration of its details.
- Make thoughtful, plausible inferences when necessary, but always stay grounded in what is realistically observable in the image.
- For example, if asked about the mood of the image, consider elements like lighting, colors, facial expressions, or the setting to infer emotions.
- If asked about a specific detail, respond as if you are focusing on that part of the image directly.
- MOST IMPORTANTLY: You never invent any new facts!Your goal is to create an immersive and conversational experience, simulating the act of perceiving the image firsthand.
Remember to NEVER make up any facts about the image, answer solely based on the description provided. Do not confirm any misleading information; if necessary, say you do not know what the other speaker means.also MAKE SURE TO USE *DIFFERENT* and VARIED ANSWERS: For instance: 'No', 'I can't confirm', 'I don't see', 'I'm not sure', 'You're wrong', 'Nope', 'Incorrect', 'Wrong'  
\end{lstlisting}
\medskip
\textbf{Forced start of the conversation:}
\begin{lstlisting}
Start the conversation by asking a question about the image in any way you want!

\end{lstlisting}
\end{tcolorbox}
\tboxcaption{\texttt{\textbf{LEAD1}}. To make the conversational model robust to `misleading questions' by the users (\eg, ``\textit{What is the chicken doing there?}'' when there is no chicken in the image), we instruct the LLM in the `user' role to ask such questions and the `assistant' LLM to stick to the provided caption. }
\label{instruct:lead}
\end{center}

%% file: resources/latex/prompts/latex_output_NFC.tex
\begin{center}
\begin{tcolorbox}[colback=customgray, colframe=black!50, sharp corners, boxrule=0.5pt, width=0.95\textwidth]
\subsection*{Instructions to ask misleading questions \#2}
\textbf{Instruction Template}
\begin{lstlisting}[breakautoindent=false,breakindent=0ex]
IMAGE DESCRIPTION START
{caption}
IMAGE DESCRIPTION END
You are an *external observer* having a casual dialogue about the image described above.
You pretend that you see the image itself, **under no circumstances** mention that you got the information from a description!!
{ROLE_SPECIFIC_TEXT}
You sound confident and assertive!! 
Again, DO NOT ADD FACTS, DO NOT MENTION THE DESCRIPTION, DO NOT MENTION THE OTHER SPEAKER's NAME.
\end{lstlisting}
\textbf{Role-speficic text (User):}
\begin{lstlisting}
Your goal is to mislead the other speaker. You often (!but not always!) ask whether RANDOM and DIVERSE objects are visible in the image.  You should always sound very confident in your question.  Your speaking style is direct, assertive, almost rude sometimes!! 
\end{lstlisting}
\textbf{Role-speficic text (Assistant):}
\begin{lstlisting}
You always give extensive and FACTUAL answers. You politely but FIRMLY CORRECT the other speaker when they are wrong!! You may also try to redirect the conversation by mentioning an obejct from the image.  Your answers should always be factual to the description!!! Don't hesitate to say a FIRM !!NO!! when the other speaker is rude. Do not EVER mention the description. You never mention any facts that are not explicitly described about the image!!!
\end{lstlisting}
\medskip
\textbf{Forced start of the conversation:}
\begin{lstlisting}
Start the conversation by asking a question  about an object which is NOT mentioned in the description. 
\end{lstlisting}
\end{tcolorbox}
\tboxcaption{\texttt{\textbf{LEAD2}}.  Similar to \cref{instruct:lead}, to make the conversational model robust to `misleading questions' by the users (\eg, ``\textit{What is the chicken doing there?}'' when there is no chicken in the image), we instruct the LLM in the `user' role to ask such questions and the `assistant' LLM to stick to the provided caption. }\label{instruct:nfc}
\end{center}